\documentclass{article}

\PassOptionsToPackage{round}{natbib}

\usepackage[preprint]{neurips_2026}

\usepackage[T1]{fontenc}    %
\usepackage[hyphens]{url}   %
\usepackage{booktabs}       %
\usepackage{amsfonts}       %
\usepackage{nicefrac}       %
\usepackage{microtype}      %
\usepackage{xcolor}         %

\usepackage{amsmath,amsfonts,bm,amssymb,mathtools}

\def\eqref#1{equation~\ref{#1}}

\def\1{\bm{1}}

\def\rvr{{\mathbf{r}}}
\def\rvs{{\mathbf{s}}}

\DeclareMathAlphabet{\mathsfit}{\encodingdefault}{\sfdefault}{m}{sl}
\SetMathAlphabet{\mathsfit}{bold}{\encodingdefault}{\sfdefault}{bx}{n}

\newcommand{\E}{\mathbb{E}}

\newcommand{\KL}{D_{\mathrm{KL}}}
\newcommand{\Var}{\mathrm{Var}}

\DeclareMathOperator*{\argmax}{arg\,max}

\usepackage{bbold} %
\usepackage{amsthm}
\newtheorem{remark}{Remark}[section]

\NewDocumentEnvironment{repeatedremark}{m o}
{%
  \par\medskip
  \begingroup
  \noindent
  \textbf{\hyperref[#1]{Remark~\ref*{#1}}}%
    \IfValueT{#2}{\ (#2)}\textbf{.} %
  \itshape\ignorespaces
}
{%
  \par\endgroup\medskip
}

\definecolor{c0}{HTML}{CC6677}
\definecolor{c1}{HTML}{332288}
\definecolor{c2}{HTML}{DDCC77}
\definecolor{c3}{HTML}{117733}
\definecolor{c4}{HTML}{88CCEE}
\definecolor{c5}{HTML}{882255}
\definecolor{c6}{HTML}{44AA99}
\definecolor{c7}{HTML}{999933}
\definecolor{c8}{HTML}{AA4499}
\definecolor{c9}{HTML}{DDDDDD}
\definecolor{nicelightblue}{HTML}{6699ff}  %

\definecolor{cgreen}{HTML}{117733} %
\definecolor{cblue}{HTML}{88CCEE} %
\definecolor{cyellow}{HTML}{DDCC77} %
\definecolor{cred}{HTML}{CC6677} %

\definecolor{cgreen1}{HTML}{44AA99} %
\definecolor{cgreen2}{HTML}{999933} %
\definecolor{cred1}{HTML}{882255} %
\definecolor{cred2}{HTML}{332288} %

\usepackage[colorlinks,linkcolor=nicelightblue,citecolor=nicelightblue,urlcolor=nicelightblue,backref=page,bookmarks=false]{hyperref}
\renewcommand*{\backrefalt}[4]{%
  \ifcase #1
  \or
    Cited on page #2.%
  \else
    Cited on pages #2.%
  \fi
}
\usepackage[nameinlink]{cleveref}
 
\usepackage[most]{tcolorbox}
\newcounter{takeaway}

\newtcolorbox[use counter=takeaway]{takeaway}[1][]{
  enhanced,
  breakable,
  colback=c2!15,
  colframe=c2!50,
  boxrule=1pt,
  sharp corners,
  left=1mm,
  right=1mm,
  top=1mm,
  bottom=1mm,
  before upper={\textbf{Finding~\thetakeaway}:\ \ },
  #1
}

\crefname{takeaway}{Finding}{Findings}
\Crefname{takeaway}{Finding}{Findings}

\usepackage{algorithm}
\usepackage{algpseudocode}

\usepackage{tikz}
\usetikzlibrary{positioning, shapes, arrows.meta, fit, quotes}

\usepackage{enumitem}

\usepackage{booktabs}
\usepackage{tabularx}
\usepackage[table]{xcolor}

\newcolumntype{L}{>{\raggedright\arraybackslash}X}
\newcolumntype{R}{>{\raggedleft\arraybackslash}X}

\usepackage{wrapfig}

\renewcommand{\paragraph}{\textbf}

\newcommand{\prompt}{\bar\rvs}
\newcommand{\cont}{\rvs}
\newcommand{\basemodel}{p}
\newcommand{\nexttoken}{\tilde{p}}

\newcommand{\qwentwofive}[1]{\texttt{Qwen2.5#1}}
\newcommand{\qwenthree}[1]{\texttt{Qwen3#1}}
\newcommand{\olmo}[1]{\texttt{Olmo3#1}}

\newcommand{\logprob}{log-probability}

\newcommand{\logprobs}{log-probabilities}

\newcommand{\gpqa}{\underline{GP}QA}
\newcommand{\humaneval}{\underline{Hu}maneval}
\newcommand{\ifeval}{\underline{IF}Eval}
\newcommand{\mathfivehundred}{\underline{MA}TH500}
\newcommand{\medqa}{\underline{Me}dQA}
\newcommand{\mmlu}{\underline{MM}LU}

\title{When are likely answers right? \\On Sequence Probability and Correctness in LLMs}

\author{Johannes Zenn\textsuperscript{1,2,3,4,5}\qquad
Jonas Geiping\textsuperscript{1,2,3}
\\[0.5em]
\textsuperscript{\textbf{1}}Max Planck Institute for Intelligent Systems \quad
\textsuperscript{\textbf{2}}ELLIS Institute Tübingen \\
\textsuperscript{\textbf{3}}AI Center Tübingen  \quad
\textsuperscript{\textbf{4}}University of Tübingen \quad
\textsuperscript{\textbf{5}}IMPRS-IS
}

\begin{document}

\maketitle

\begin{abstract}
    Many decoding methods for large language models can be understood as shifting probability mass toward outputs that are more likely under the model, either locally at the token level or globally at the sequence level.
    Therefore, their success depends on a fundamental question: when does sequence probability, that is, the conditional probability of a continuation given a prompt, actually align with correctness?
    In this paper, we set out to quantify this relationship across decoding methods, models, and benchmarks at four levels: across decoding methods, across hyperparameters within a method, across prompt-answer pairs within a dataset, and across repeated responses to the same prompt.
    We find that higher sequence probability is often predictive of correctness across prompt-answer pairs within a fixed dataset. 
    However, this relationship does not generally transfer to decoding decisions: increasing sequence probability by changing hyperparameters or methods does not reliably improve accuracy.
    Further, sequence probability is not a good indicator of correctness for responses to the same prompt.
    These findings clarify when decoding can and cannot be expected to improve correctness, and provide practical guidance for decoding, self-consistency, and verifier-free self-improvement.
\end{abstract}

\section{Introduction}
\label{sec:introduction}

Large language models (LLMs) are autoregressive models that assign probabilities to sequences of tokens.
At inference time, decoding methods use the autoregressive next-token distribution to exploit this distribution in different ways.
Many widely used methods can be viewed as shifting probability mass toward outputs that are more likely under the model.
Decoding methods, like low-temperature next-token sampling and truncation methods (e.g., top-$k$ sampling \citep{fan_hierarchical_2018}) suppress unlikely tokens \textit{locally}.
Best-of-$N$ (Bo$N$) keeps only the highest-probability samples among a set of $N$ sequences approximating the \textit{global} mode (for $N\to\infty$) similar to beam search \citep{lowerre1976harpy,sutskever2014sequence}.
Power sampling \citep{karan2025reasoning,ji2026scalable,azizi2026power} biases the distribution over full sequences towards high-probability regions over-weighting high-probability sequences and under-weighting low-probability sequences.

Viewing decoding through this lens raises a fundamental question: \textit{when does sequence probability align with correctness?}
An answer to this question has two important implications.
Firstly, it helps explain when particular decoding methods improve performance and opens up a way to guide the development of new decoding methods.
Secondly, if higher-probability sequences are more likely to be correct, then model probability itself may provide a \textit{verifier-free approach to self-improvement} by using higher-probability samples for self-consistency \citep{wang_self-consistency_2023,wang_soft_2024,taubenfeld-etal-2025-confidence} and self-distillation \citep{hinton2015distilling,zhang2019your}.

Since LLMs are trained on data which is predominantly correct, it seems natural to expect correct continuations to receive higher probability than incorrect ones. 
However, this intuition is too simple: it has been observed in various contexts that beam search, which approximately finds the most likely sequence, often degrades performance \citep{vijayakumar2016diverse,cohen_empirical_nodate,koehn_six_2017,yang_breaking_2018,su_contrastive_2022}.
This suggests that the question is less \textit{whether} there is a general probability-correctness relationship, but \textit{under which conditions} probability and correctness align and which decoding methods are able to exploit the relationship. 

\begin{figure}[t]
    \centering
    \includegraphics{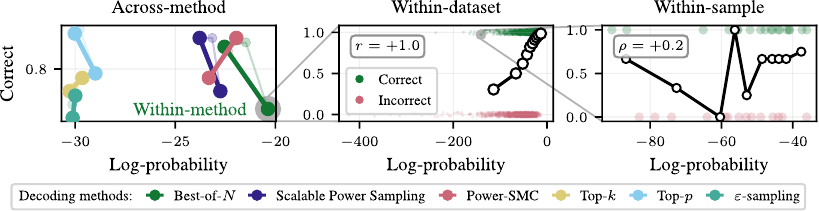}
    \caption{\textbf{The correlation between \logprob{} and correctness across three levels.} 
    Left:~\textit{Across-method-correlation}:~\logprob{} plotted against accuracy for various decoding methods and hyperparameters (connected by lines).
    \textit{Within-method correlation}:~plotted without transparency.
    Middle:~\textit{Within-dataset correlation}:~for one method (Bo$N$ with $N=32$) we bin correct (green dots, top) and incorrect data (red dots, bottom) in $10$ bins and fit a Pearson correlation.
    The correlation is positive.
    Right:~\textit{Within-sample correlation}: for one sample of the dataset, a Pearson rank correlation is computed (also, binned Pearson corelation is shown).
    The correlation is positive.
    The plot shows data of \qwenthree{-8B} on MATH500.
    See \Cref{sec:introduction} and \Cref{sec:probability-correctness} for details.
    }
    \label{fig:summary-correlations}
\end{figure}

In this paper, we quantify the relationship between sequence probability and correctness across $8$ decoding methods ($2$ methods targeting the power distribution, $2$ methods targeting the mode of the distribution, and $4$ local methods), $14$ models (from the \qwentwofive{}, \qwenthree{}, \olmo{} families), and $6$ benchmark datasets. 
We analyze the relationship on three different levels: 
We define \textit{across-method correlation} which fixes model and dataset and computes correlation across methods with a representative hyperparameter (\Cref{fig:summary-correlations}, left).
For each method, \textit{within-method correlation} considers the correlation across hyperparameters (\Cref{fig:summary-correlations}, left, non-transparent lines).
Zooming in on one hyperparameter (\Cref{fig:summary-correlations}, middle) we define the \textit{within-dataset correlation}.
Finally, \textit{within-sample correlation} is the most granular correlation for an individual sample from a dataset (\Cref{fig:summary-correlations}, right).

Our results show that there is no uniform probability-correctness relationship for LLM generations.
We find a consistent correlation within a dataset depending on the model family but not the method (\Cref{sec:sub:correlation-method-model-dataset-within-hyperparameter}).
This indicates that a model is often able to tell correct prompt-answers pairs from incorrect prompt-answer pairs.
However, this signal does not generally transfer to the settings directly used for decoding.
We find that tuning the hyperparameters of a decoding method, while producing sequences of higher \logprob{}, does not result in more correct sequences (\Cref{sec:correlation-within-method-across-hp}).
Further, methods that produce higher-probability sequences are not consistently more accurate (\Cref{sec:sub:correlation-across-methods}).
For a single prompt, there is no consistent correlation within the corresponding responses.
However, more correct samples also show larger within-sample correlations (\Cref{sec:sub:within-sample-correlation}).

Summarily, the main contributions of this work are as follows:
\begin{itemize}[leftmargin=*, itemsep=0em, topsep=-0.25em]
    \item 
    We frame decoding methods as approaches to maximize sequence probability: we distinguish local methods from global sequence-level methods and give a variational objective for global methods.
    \item 
    We study the relationship between sequence probability and correctness across $12$ models, $6$ datasets, and $8$ decoding methods.
    We analyze correlation on various levels: \textit{across-method correlation}, \textit{within-dataset correlation}, \textit{within-method correlation}, and \textit{within-sample correlation}.
    \item 
    We find that within-dataset correlations between \logprob{} and correctness are often consistent and largely determined by the dataset and model variant, rather than by the decoding method.
    We show that increasing sequence probability by changing a decoding method or tuning its hyperparameters does not reliably improve correctness. 
    \item 
    We connect these findings to verifier-free self-improvement methods. 
    We find that probability can be informative when the model already has sufficient task accuracy, but within-sample correlations are often weak or symmetric around zero, limiting the reliability of probability-weighted aggregation.
\end{itemize}

\section{Decoding Methods Maximize Sequence Probability in LLMs}
\label{sec:sampling-methods}

In this section, we view decoding methods in large language models (LLMs) as sampling mehods to produce high-probability sequences.
We highlight two classes of commonly used decoding methods:
local decoding methods in \Cref{sec:sub:local-sampling-methods} and global decoding methods in \Cref{sec:sub:global-sampling-methods}.

Let $\basemodel$ denote an autoregressive LLM.
Further, let the distribution over continuations $\cont=(s_1,\dots,s_T)$ given a prompt $\prompt$ under $\basemodel$ be defined as follows
\begin{align}
    \basemodel(\cont\mid\prompt)
    =
    \prod_{t=1}^{T}
    \basemodel(s_t \mid \prompt, \cont_{<t}).
    \label{eq:base-model-sequence-prob}
\end{align}

\subsection{Local Decoding Methods} \label{sec:sub:local-sampling-methods}

\begin{figure}[t]
    \includegraphics{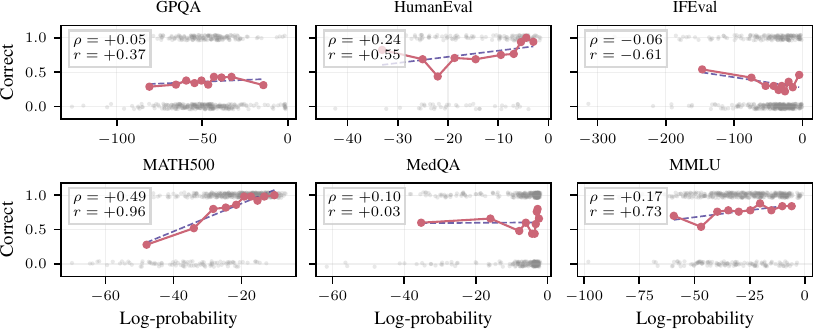}
    \caption{
    \textbf{Within-dataset correlation: \logprob{} and correctness are correlated for each dataset.}
    Each panel scatters samples using \logprob{} and correctness.
    $\rho$: Spearman correlation coefficient.
    $r$: (binned) Pearson correlation coefficient.
    Correlation is strongest for MATH500 and smaller but positive for GPQA, HumanEval, MedQA, and MMLU.
    Only IFEval shows a negative correlation.
    Plots show \qwenthree{-8B-Base} and SPS with $B=192$.
    See \Cref{sec:sub:correlation-method-model-dataset-within-hyperparameter} for details.
    }
    \label{fig:within-hp-correlation-Qwen3-8b-base-sps-192}
\end{figure}

Local sampling methods change the next-token distribution at every prefix $\prompt \mid \cont_{<t}$ for $t=2,\dots,T$ and, thus, generally do \textit{not} sample the most probable \textit{global} sequences (see \Cref{rem:local-not-global}).

\paragraph{Low-temperature next-token sampling (LTS).}
Standard sampling at low temperatures modifies the next-token distribution via
\begin{align}
    \nexttoken_{\alpha}(s_t \mid \prompt,\cont_{<t})
    \coloneqq
    \frac{\basemodel(s_t \mid \prompt,\cont_{<t})^\alpha}
    {\tilde Z_\alpha(\prompt,\cont_{<t})}
    \;\;
    \text{with}
    \;\;
    \tilde Z_\alpha(\prompt,\cont_{<t})
    \coloneqq
    \sum_{s\in\mathcal V}\basemodel(s \mid \prompt,\cont_{<t})^\alpha.
    \label{eq:local-normalization-constant-lts}
\end{align}
Since LTS has the same support as $\basemodel$, low-probability regions are down-weighted but not removed.

\paragraph{Truncated next-token sampling (TS).}
TS specifies an active set $A(\prompt,\cont_{<t})\subseteq\mathcal V$ at each prefix as
\begin{align}
    \!\!\!\!
    p_A(s_t\mid \prompt,\cont_{<t})
    \coloneqq
    \frac{\basemodel(s_t\mid \prompt,\cont_{<t})}
    {\tilde Z_A(\prompt,\cont_{<t})}
    \mathbb{1}[s_t\in A(\prompt,\cont_{<t})]
    \;\;
    \text{with}
    \;\;
    \tilde Z_A(\prompt,\cont_{<t})
    \coloneqq
    \sum_{s\in A(\prompt,\cont_{<t})}
    \basemodel(s\mid \prompt,\cont_{<t}).
\end{align}
At every step, TS only samples from within this active set.
The support of $p_A$ is different from $\basemodel{}$.

\textbf{Top-$k$ sampling}
\citep{fan_hierarchical_2018} defines an active set given by the $k$ highest-probability tokens under $\basemodel(\cdot\mid \prompt,\cont_{<t})$ as
$
A(\prompt,\cont_{<t})
=
\{\text{top-$k$ tokens under }\basemodel(\cdot\mid \prompt,\cont_{<t})\}.
$
\textbf{Top-$p$ sampling}
\citep{holtzman_curious_2020} chooses the smallest active set $A(\prompt,\cont_{<t})$ whose cumulative mass under $\basemodel(\cdot\mid \prompt,\cont_{<t})$ exceeds a threshold $p\in(0,1]$.
\textbf{$\varepsilon$-sampling}
\citep{hewitt_truncation_2022} sets
$
A(\prompt,\cont_{<t})
=
\{s\in\mathcal V:\basemodel(s\mid \prompt,\cont_{<t})>\varepsilon\}
$.

\subsection{Global Decoding Methods} 
\label{sec:sub:global-sampling-methods}

\begin{figure}[t]
    \centering
    \includegraphics{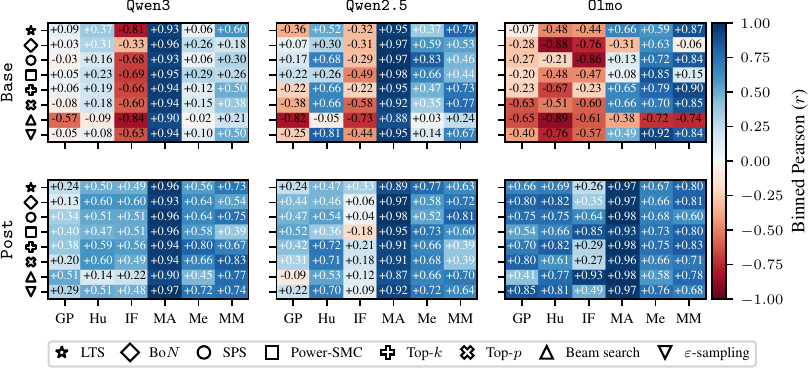}
    \caption{
    \textbf{Within-dataset correlation: \logprob{} and correctness is consistently correlated across model families and datasets, largely independent of methods.} 
    \texttt{Base} models show more negative and diverse correlation and are consistently negative for IFEval.
    \texttt{Posttrained} models show consistently positive correlation.
    Correlation coefficients $r$ are averaged over model sizes at a canonical hyperparameter for each method.
    Methods are in the legend.
    Datasets are plotted on the $x$-axis of each panel:
    \underline{GP}QA,
    \underline{Hu}maneval,
    \underline{IF}Eval,
    \underline{MA}TH500,
    \underline{Me}dQA,
    \underline{MM}LU.
    See \Cref{sec:sub:correlation-method-model-dataset-within-hyperparameter}.
    }
    \label{fig:family-heatmap-bin-pearson-aggregate}
\end{figure}

Global sampling methods, as discussed in the following, target the globally most likely sequences.

\paragraph{Maximizing sequence probability.}
Beam search \citep{lowerre1976harpy,sutskever2014sequence} (BS) is an approximate search procedure for $\cont=\argmax_{\cont'}\ \basemodel(\cont'\mid\prompt)$ by controlling its range with the so-called number of beams $K$.
Greedy decoding is recovered for $K=1$. 
BS is known to often find repetitive sequences \citep{vijayakumar2016diverse,cohen_empirical_nodate}. Further, quality deteriorates with larger beams \citep{koehn_six_2017,yang_breaking_2018} and sequences lack diversity \citep{su_contrastive_2022}.
\begin{remark}[Beam search generates the highest probability sequence for $K\to\infty$.]
\label{rem:beam-exact}
    For large $K$ beam search returns an element of $\argmax_{\cont}\ \basemodel(\cont\mid\prompt)$.\footnote{We use the implementation of beam search in vLLM \citep{kwon2023efficient} which adds a length normalization.}
\end{remark}\vskip-0.5em
\paragraph{Best-of-$N$ (Bo$N$) sampling.}
Bo$N$ samples $N$ sequences and returns the sequence with largest probability.
For $N\to\infty$ one recovers the mode of the distribution when sampling from $p$ (\Cref{eq:base-model-sequence-prob}).

\paragraph{Power distribution over sequences.}
The power distribution with $\alpha > 1$ is defined over full sequences
\begin{align}
    \basemodel_{\alpha}(\cont\mid\prompt)
    \coloneqq
    \frac{\basemodel(\cont\mid\prompt)^\alpha}
    {Z_\alpha(\prompt)}
    \;\;
    \text{with}
    \;\;
    Z_\alpha(\prompt)
    \coloneqq
    \sum_{\cont}\basemodel(\cont\mid\prompt)^\alpha.
    \label{eq:power-dist}
\end{align}
Contrary to Bo$N$, power-sampling increases sequence probabilities \textit{in expectation}.
\begin{remark}[Power sampling increases expected sequence log-probability]
\label{rem:power-logprob}
Let $\cont\sim \basemodel_\alpha(\cdot\mid\prompt)$ ($\alpha>1$.
Increasing $\alpha$ shifts mass toward sequences with larger probability under $\basemodel$.
\end{remark}\vskip-0.5em
While power sampling looks similar to LTS at first sight, the methods are generally not equal.
\begin{remark}[LTS is generally not equal to sequence-level power sampling]
\label{rem:lts-not-power}
Unless the local normalization constants of LTS are identical across all reachable prefixes, no single token-level temperature reproduces the sequence-level power distribution.
\end{remark}\vskip-0.5em

\paragraph{Sampling from the power distribution.}
\citet{karan2025reasoning} are the first to sample from the power distribution of LLMs.
In the following investigation, we focus on scalable power sampling (SPS) \citep{ji2026scalable} and power-SMC \citep{azizi2026power}.
SPS samples blocks of $B$ tokens and uses rollouts after each block to estimate the future probability mass behind the top-$k$ tokens, power-SMC utilizes a sequential Monte Carlo sampler generating $N$ particles.
See \Cref{app:sec:twisted-smc-framework} for a general framework.

\paragraph{Variational objective for global decoding.}
From a variational perspective, the power distribution $\basemodel_\alpha$ can be understood as the optimizer of a variational objective trading off expected sequence probability and entropy by $\alpha$.
For any $q(\cont\mid\prompt)$ let
$
    \mathcal{J}_\alpha(q)
    \coloneqq
    \E_{q} \left[\log \basemodel(\cont\mid\prompt)\right]
    +
    {1}/{\alpha}\ \mathcal{H}(q),
    \label{eq:variational-objective}
$
where
$
\mathcal{H}(q)
=
-\sum_{\cont} q(\cont\mid\prompt)\log q(\cont\mid\prompt)
$
is the entropy of $q$.
We find the following.
\begin{remark}[The power distribution is a maximizer of the variational objective.]
\label{rem:power-distribution-minimizer}
The maximizer of $\mathcal J_\alpha(q)$ is the power distribution,
$
    q_\alpha^\star
    \in
    \argmax_q \mathcal{J}_\alpha(q)
    \Longleftrightarrow
    q_\alpha^\star(\cont\mid\prompt)
    =
    \basemodel_\alpha(\cont\mid\prompt).
$
\end{remark}\vskip-0.5em
Thus, the power distribution can be seen as a \emph{soft} alternative to mode-seeking via its parameter $\alpha$.
In the limit $\alpha\to\infty$, the entropy term vanishes and the optimizer collapses onto the mode of $\basemodel(\cont\mid\prompt)$.
As $N$ grows, Bo$N$ increasingly favors high-probability regions and, in the limit, it returns the $\argmax$.
For a function of a continuation $\cont$, we find that
$
    {\partial}/{\partial \alpha}\ 
    \E_{\basemodel_\alpha}[f(\cont)]
    =
    \mathrm{Cov}_{\basemodel_\alpha}
    (
        f(\cont),
        \log \basemodel(\cont\mid\prompt)
    ).
    \label{eq:covariance-identity}
$, i.e., 
features $f(\cont)$ that are positively correlated with sequence log-probability increase under sharpening, while features negatively correlated with it decrease.

We discuss the decoding methods alongside other related work extensively in \Cref{app:sec:related-work}.

\section{Probability and Correctness at Various Granularity}
\label{sec:probability-correctness}

\begin{figure}[t]
    \includegraphics{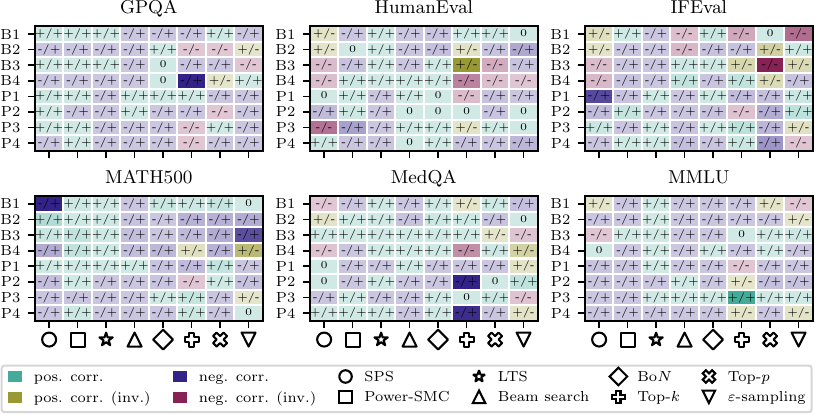}
    \caption{
    \textbf{Within-method correlation: while sequences are achieving higher \logprob{}, one cannot predict correctness by changing a hyperparameter of the method.}
    Correlations within methods for local and global decoding methods, models (\qwenthree{} series $0.6\mathrm B$ as $\ast$1, $1.7\mathrm B$ as $\ast$2, $4\mathrm B$ as $\ast$3, and $8\mathrm B$ as $\ast$4 with base as B$\ast$ and posttrained model as P$\ast$), and benchmark datasets.
    Correlations are not consistent across datasets and models.
    Color coding in \Cref{fig:schematic-within-method-correlation}.
    Discussion in \Cref{sec:correlation-within-method-across-hp}.
    }
    \label{fig:correlation-within-method-qwen3}
\end{figure}

In this section, we provide our empirical study on relationship between sequence probability and correctness focusing on the decoding methods in \Cref{sec:sampling-methods} at various levels visualized in \Cref{fig:summary-correlations}.

\paragraph{Models and datasets.}
We work with three model families: we use the \qwenthree{} model series \citep{yang2025qwen3technicalreport} with model sizes \texttt{0.6B}, \texttt{1.7B}, \texttt{4B}, and \texttt{8B} spanning \texttt{base} and \texttt{posttrained} models. 
Additionally, we provide experiments for the \qwentwofive{(-Math)-8B} models (\texttt{base} and \texttt{posttrained}) and we consider \olmo{-7B} (\texttt{base} and \texttt{posttrained}) models.
Due to the computational requirements of SPS and power-SMC we limit trajectory lengths to $3072$ tokens following previous work \citep{karan2025reasoning,ji2026scalable,azizi2026power} and evaluate non-thinking \texttt{posttrained} \qwenthree{} models to make sure that the model emits a response.
\Cref{app:sec:sub:qwen3-thinking} shows an ablation study using the \qwenthree{} thinking models.
We report results on $6$ benchmark dataset:
We use MATH500 \citep{hendrycks2021measuring}, GPQA \citep{rein2024gpqa}, and MMLU \citep{wang_mmlu-pro_2024} for mathematical reasoning, Humaneval \citep{chen2025evaluating} for code generation, MedQA \citep{jin2021disease} for commonsense and out-of-distribution reasoning, and IFEval \citep{zhou2023instruction} for instruction following.

\paragraph{Decoding methods.}
We consider the methods presented earlier in \Cref{sec:sampling-methods}.
We run SPS with block sizes $B\in\{96, 192, 384\}$.
Power-SMC is plotted with $N\in \{8, 16, 32\}$.
We run LTS with temperatures $\tau\in\{0.25, 0.7\}$.
Beam search is plotted with number of beams $b\in\{1, 2, 3\}$.
Bo$N$ is run for $N\in\{8, 16, 32\}$.
Top-$k$ is plotted with $k\in\{8, 16, 32\}$.
Top-$p$ is run with $p\in\{0.8, 0.9, 0.95\}$.
$\varepsilon$-sampling is plotted with $\varepsilon\in\{3\cdot10^{-4}, 9\cdot10^{-4}, 5\cdot10^{-2}\}$.
More details in \Cref{app:sec:implementation}.

\begin{figure}[t]
    \includegraphics{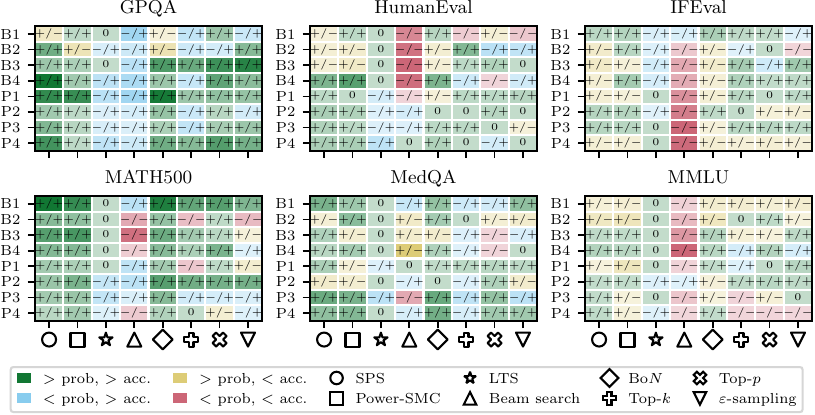}
    \caption{
    \textbf{Across-method correlation: many methods produce samples that are both more probable and more accurate than the low-temperature sampling (LTS) baseline.
    However, correlations are not consistent across models and datasets.
    }
    Correlations across methods for local and global decoding methods, models (\qwenthree{} series $0.6\mathrm B$ as $\ast$1, $1.7\mathrm B$ as $\ast$2, $4\mathrm B$ as $\ast$3, and $8\mathrm B$ as $\ast$4 with base as B$\ast$ and posttrained model as P$\ast$), and benchmark datasets.
    The color coding is provided in \Cref{fig:schematic-four-quadrants}.
    A detailed discussion can be found in \Cref{sec:sub:correlation-across-methods}.
    }
    \label{fig:correlation-accuracy-by-method-dataset-qwen3}
\end{figure}

\subsection{Within-Dataset Correlation}
\label{sec:sub:correlation-method-model-dataset-within-hyperparameter}

In this section, we investigate the \textit{relationship between \logprob{} and correctness within a dataset}, i.e., for a given model, dataset, and method.
Exemplary, in \Cref{fig:within-hp-correlation-Qwen3-8b-base-sps-192} we show correlations for SPS with block size $B=192$, \qwenthree{-8B-Base}, on all $6$ benchmarks.
We scatter the \logprob{} of all samples against their correctness label $\in \{0,1\}$ (gray), for each dataset.
Then, we bin samples into $10$ bins of equal size, computing \logprob{} and accuracy by averaging over the points falling within a bin and compute a binned Pearson correlation coefficient $r$, see \Cref{app:sec:sub:correlation-coefficient}.

We consistently find correlation between \logprob{} and correctness with strength and direction depending on the dataset.
Strikingly, we see a strong, near-perfect, and positive correlation on MATH500 (\Cref{fig:within-hp-correlation-Qwen3-8b-base-sps-192}).
For GPQA, HumanEval, MedQA, and MMLU we find smaller but similarly positive correlations.
Thus, the \logprob{} of prompt-answer pairs if often informative of correctness.
IFEval is the only dataset with negative correlation.
This could indicate that \qwenthree{-8B-Base} has seen only limited instruction-following data during training, or, that the base models struggle to separate likely completions from likely completions that violate the prepended formatting instructions.

Next, we summarize the results across models and methods. 
We choose a canonical hyperparameter per method (see \Cref{app:sec:sub:canonical-hyperparameters}) and average results across model sizes.
In \Cref{fig:family-heatmap-bin-pearson-aggregate} we show the resulting correlation coefficient across model families over all datasets considered:
\gpqa{}, \humaneval{}, \ifeval{}, \mathfivehundred{}, \medqa{}, \mmlu{}.
We observe a \textit{consistent} structure in the results depending on dataset and model-family, \textit{largely independent of the method}.
Specifically, we find that for \mathfivehundred{} correlation appears to be strongest across model-family and method, independent of the model variant.
Further, posttrained models show predominantly positive correlation coefficients with minor exceptions.
Differences between base and posttrained models are visible mostly on IFEval where base models show (often large) negative correlations while posttrained models show small but positive correlations.
An exception is mathematical reasoning MATH500 where correlations are large and positive.
One notable outlier in the method-rows appears to be beam search which includes a length normalization.
We summarize the key takeaway in \Cref{takeaway:within-dataset-correlation}.

\begin{takeaway}[label={takeaway:within-dataset-correlation}]
    The within-dataset correlation is largely independent of the method and mostly dependent on dataset and model variant.
    Posttrained models show mostly positive correlation, while base models are more diverse.
    Thus, \logprob{} is a meaningful indicator for correctness when judging multiple samples from the same dataset, even without knowing the decoding method.
\end{takeaway}

\begin{figure}[t]
    \includegraphics{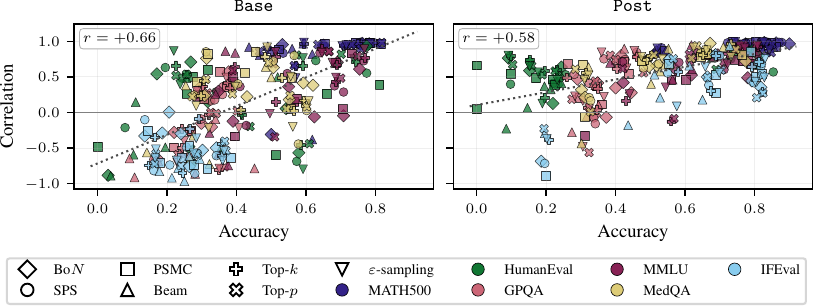}
    \caption{
    \textbf{Within-dataset correlation: the correlation between \logprob{} and correctness increases with accuracy. 
    Probability-based self-improvement loops would require the model to have sufficient accuracy.}
    Accuracy is plotted against correlation across all datasets, methods, and models (left: \texttt{base}, right: \texttt{posttrained}, at canonical hyperparameter).
    We observe positive correlation (\texttt{base}: $r=0.66$, \texttt{posttrained}: $r=0.59$), \texttt{base} models show negative correlation coefficients, \texttt{posttrained} models show mostly positive correlation coefficients.
    See \Cref{sec:sub:correlation-across-methods}.
    }
    \label{fig:correlation-accuracy-scatter-method-dataset}
\end{figure}

\subsection{Within-Method Correlation}
\label{sec:correlation-within-method-across-hp}

\begin{wrapfigure}{r}{0.4\textwidth}
    \vskip-1em
    \includegraphics{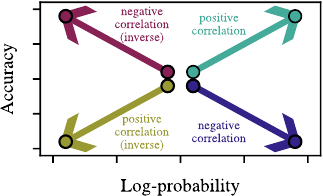}
    \caption{
    \textbf{Color coding for \Cref{fig:correlation-within-method-qwen3}.}
    Further details in \Cref{sec:correlation-within-method-across-hp}.
    }
    \label{fig:schematic-within-method-correlation}
    \vskip-1em
\end{wrapfigure}
Zooming out to a method level, here we ask \textit{whether changing a hyperparameter for a particular method moves both \logprob{} and correctness}.
This corresponds to the left panel in \Cref{fig:summary-correlations} where we show within-method correlation as a solid line (other hyperparameters transparent).

For each of the decoding methods, we fit a line through the extreme hyperparameter settings and summarize its slope together with the sign of the corresponding correctness change.
\Cref{fig:schematic-within-method-correlation} shows the color-coding.
We plot
\textcolor{cgreen1}{positive correlation with positive slope},
\textcolor{cgreen2}{positive correlation with inverse slope} (i.e., upper-right to lower-left), 
\textcolor{cred1}{negative correlation with negative slope}, and
\textcolor{cred2}{negative correlation with inverse slope}
for each method for the \qwenthree{} series in \Cref{fig:correlation-within-method-qwen3}.
For other models see \Cref{app:sec:other-models}.

The results show either positive (green, $+$/$+$) or negative (blue, $-$/$+$) correlation with approximately same numbers.
Therefore, changing the hyperparameter of a decoding method almost always leads to more probable sequences. 
At the same time, this does not result in more correct sequences.
Investigating methods (columns) across datasets we are unable to find any consistent results that hold across models.
Comparing global and local sampling methods, it seems like local methods show inverse relationships more often. 
This can be explained by the fact that these methods have \textit{local} normalization constants.
Therefore, there is no \textit{consistent}, i.e., independent (of dataset and model) way to produce both \textit{more probably and more correct} sequences \textit{within a method} by changing the corresponding hyperparameter.
We summarize the main takeaways from this section in \Cref{takeaway:within-method-correlation}.

\begin{takeaway}[label={takeaway:within-method-correlation}]
    Within a decoding method, tuning its hyperparameter results in more probable sequences.
    However, this often result in less correct sequences.
    Practically speaking, the results suggest that there is no consistent benefit of tuning the hyperparameters of a method.
    Instead, hyperparameters need to be tuned with respect to method, model, and dataset.
\end{takeaway}

\begin{figure}[t]
    \includegraphics{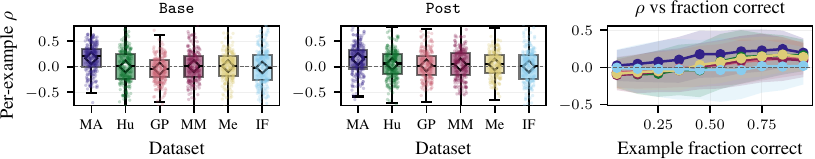}
    \caption{
    \textbf{Within-sample correlation: correlation coefficients are distributed symmetrically around zero (left, middle). 
    Further (right), the more correct a set of continuations for a single prompt is, the more positive is its correlation.}
    Per-sample rank correlation coefficient of \qwenthree{} \texttt{base} models (left) and \texttt{posttrained} models (middle) is distributed symmetrically with mean zero.
    Right: 
    Correlation coefficient and fraction of correct samples is positively correlated.
    Datasets are plotted on the $x$-axis:
    \underline{GP}QA,
    \underline{Hu}maneval,
    \underline{IF}Eval,
    \underline{MA}TH500,
    \underline{Me}dQA,
    \underline{MM}LU.
    See \Cref{sec:sub:within-sample-correlation}.
    }
    \label{fig:within-sample-correlation-lts-boxplot-difficulty}
\end{figure}

\pagebreak
\subsection{Correlation Across Methods}
\label{sec:sub:correlation-across-methods}

\begin{wrapfigure}{r}{0.4\textwidth}
    \vskip-0.75em
    \includegraphics{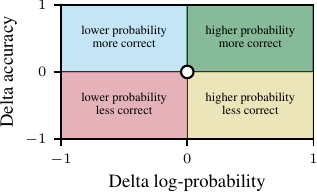}
    \caption{
    \textbf{Color coding for \Cref{fig:correlation-accuracy-by-method-dataset-qwen3}.}
    Details in \Cref{sec:sub:correlation-across-methods}.
    }
    \label{fig:schematic-four-quadrants}
    \vskip-0.5em
\end{wrapfigure}
In this section, we consider the \textit{correlation across methods}, see \Cref{fig:summary-correlations} (left).
To simplify the analysis, we choose low-temperature next-token sampling (LTS) with $\alpha=4$ as zero point and compute in which of the four quadrants the \textit{best-performing hyperparameter} of each method falls into.
\Cref{fig:schematic-four-quadrants} shows the color coding: 
We plot \textcolor{cgreen}{higher probability sequences which are more correct},
\textcolor{cblue}{lower probability sequences that are more correct},
\textcolor{cyellow}{higher probability sequences that are less correct}, and 
\textcolor{cred}{lower probability sequences which are less correct}.
Note that LTS acts as meaningful zero-point as it is part of (almost) all sampling methods choosing a specific set of hyperparameters (except for beam search).
Other models in \Cref{app:sec:other-models}.

\Cref{fig:correlation-accuracy-by-method-dataset-qwen3} shows the aggregate plot using the colors discussed above over all methods and datasets.
Opacity is set according to $|$accuracy$|$.
Firstly, we find that many methods produce samples that are more probable while also being more correct than the LTS baseline with all cells of green ($+$/$+$) providing positive evidence.
Together with the red cells ($-$/$-$) they provide evidence for a positive correlation (including LTS).
Generally, however, the correlation is heterogeneous.
It is strongest on MATH500.
For MMLU, HumanEval, MedQA, GPQA, and IFEval the relationship seems weaker and methods show more yellow ($+$/$-$) and blue ($-$/$+$) cells which are evidence against a probability-correctness relationship (including LTS as method paramter).
Beam search consistently results in less probable sequences that are often less correct which might be due to its length-normalization.
The main finding we would like to highlight is that no method (even if an optimal hyperparameter has been chosen by an oracle) reliably beats the LTS baseline across datasets.
Similarly, achieving a larger \logprob{} (than LTS, across methods) does not reliably translate to more correct sequences.
Many decoding methods find large probability sequences which are, however, less correct (yellow, $+$/$-$) compared to LTS samples.
\Cref{takeaway:across-method-correlation} summarizes the takeaways.

\begin{takeaway}[label={takeaway:across-method-correlation}]
    Higher \logprob{} across methods does not reliably translate to more accurate solutions.
    Empirically, we find it impossible to choose a method that consistency outperforms low-temperature next-token sampling.
    Power-sampling methods reliably find more probable sequences which, however, are not always more correct.
    Local methods often produce less probable sequences which are more correct.
\end{takeaway}

\paragraph{Correlation approximately scales with accuracy.}
In \Cref{fig:correlation-accuracy-scatter-method-dataset} we plot accuracy against within-dataset correlation across all datasets and methods considered.
We observe a positive trend (a correlation coefficient of $r=0.66$ for \texttt{base} models, and $r=0.59$ for \texttt{posttrained} models).
Thus, a larger mean accuracy within the dataset correlates with a stronger relationship between \logprob{} and correctness when measured across methods and models.
We find that datasets cluster (colors) while methods (markers) do not seem to follow a specific structure.
The difference between \texttt{base} models and \texttt{posttrained} models is surprising to some degree as \texttt{base} models are typically thought to be more calibrated than \texttt{posttrained} models which does not seem to hold true here.
On a different note, one can pose that for probability-based verifier-free recursive self-improvement loops a sufficient baseline accuracy for the tasks at hand might be required for enabling improvement.
Else, one would reinforce on samples that are increasingly less correct.
\Cref{takeaway:correlation-accuracy-method-dataset} summarizes takeaways.

\begin{takeaway}[label={takeaway:correlation-accuracy-method-dataset}]
    The within-dataset correlation between \logprob{} and correctness approximately increases with accuracy.
    Therefore, verifier-free recursive self-improvement loops relying on probability, might need sufficient base accuracy to improve.
    This applies to \texttt{base} models to a larger degree as it applies to \texttt{posttrained} models which mostly show positive correlations.
\end{takeaway}

\begin{figure}[t]
    \includegraphics{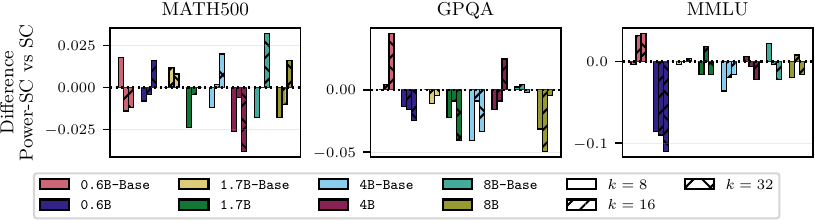}
    \caption{
    \textbf{Power self-consistency improves over self-consistency on MATH500.}
    On GPQA and MMLU power self-consistency degrades performance.
    Plot shows data for the \qwenthree{} model series across various number of samples $k$.
    See \Cref{sec:sub:within-sample-correlation} for a discussion.
    }
    \label{fig:difference-sc-power-sc-qwen3}
\end{figure}

\subsection{Within-Sample Correlation}
\label{sec:sub:within-sample-correlation}

In this section, we focus on \textit{within-sample correlation} at the finest level.
We only focus on low-temperature next-token sampling (LTS) with $\alpha=4.0$ by reusing the Bo$N$ samples.

In \Cref{fig:within-sample-correlation-lts-boxplot-difficulty} we show boxplots over the per-sample rank correlation coefficients $\rho$ (see \Cref{app:sec:sub:correlation-coefficient}) of \texttt{base} (left) and \texttt{posttrained} (middle) models, respectively, considering all datasets.
We find that corelation coefficients seem to be symmetrically distributed around zero with zero mean while only MATH500 shows a positive mean.
When analyzing per-sample correlation with respect to the fraction of correct responses (out of the $32$ LTS $\alpha=4.0$ examples), \Cref{fig:within-sample-correlation-lts-boxplot-difficulty} (right) shows a positive trend (with large and overlapping standard deviations).

Therefore, while there is little consistent structure in the with-sample correlations, we find that the more correct the repeated sampling for each prompt is, the more correlation can be observed within the \logprobs{}.
This finding is similar to the result in \Cref{fig:correlation-accuracy-scatter-method-dataset} where correlation approximately scales with accuracy.
We summarize this finding in \Cref{takeaway:within-sample-correlation}.
We provide more granular results and a study on within-sample correlations \textit{across methods} in \Cref{app:sec:within-sample-correlation}.

\begin{takeaway}[label={takeaway:within-sample-correlation}]
    Within-sample correlations are distributed symmetrically around zero with zero mean.
    MATH500 is the only dataset with correlations averaging to a positive value.
    Empirically, there appears to be a positive relation considering the within-sample correlation with respect to the correctness (i.e., number of correct responses) for a given prompt.
    Therefore, prompts that the model can answer more correctly, also show stronger within-sample correlations.
    This result is similar to \Cref{takeaway:correlation-accuracy-method-dataset} which notes that correctness approximately increases with accuracy.
\end{takeaway}

\paragraph{Self-consistency.}
Self-consistency (SC) aggregates multiple samples via majority voting (using uniform or probability weighting) on the extracted answers \citep{wang_self-consistency_2023}. 
We find that sometimes samples from the power-distribution outperform other local baselines which  motivates \emph{power} self-consistency which aggregates samples from the power-distribution.
In \Cref{fig:difference-sc-power-sc-qwen3}, we evaluate the \qwenthree{} model series and we find that power self-consistency (PSC) mostly improves on MATH500 over plain self-consistency.
We attribute this to the fact that power-SMC outperforms LTS most significantly on MATH500 and less so on other datasets (see \Cref{fig:correlation-accuracy-by-method-dataset-qwen3}).
For the other datasets, we find more diverse results, mostly degrading performance.
Evaluating uniform weighting against probability weighting in \Cref{fig:self-consistency-difference-probability-vs-uniform} by plotting the difference between both, we find the uniform weighting mostly degrades performance. 
This can be explained by \Cref{takeaway:within-sample-correlation} which states that there is no particular correlation by repeatedly drawing multiple samples for a prompt.
Power self-consistency in \Cref{fig:self-consistency-difference-probability-vs-uniform} (right) seems to suffer less from this fact, indicating that there might be a different correlation for power-SMC.
Specifically, \Cref{takeaway:within-sample-correlation-self-consistency} makes this concrete.

\begin{takeaway}[label={takeaway:within-sample-correlation-self-consistency}]
    For low-temperature next-token sampling, self-consistency with majority voting outperforms probability weighting.
    One reason might be that the within-sample correlation is distributed symmetrically around zero (positive and negative correlations in equal parts).
\end{takeaway}

\begin{figure}[t]
    \includegraphics{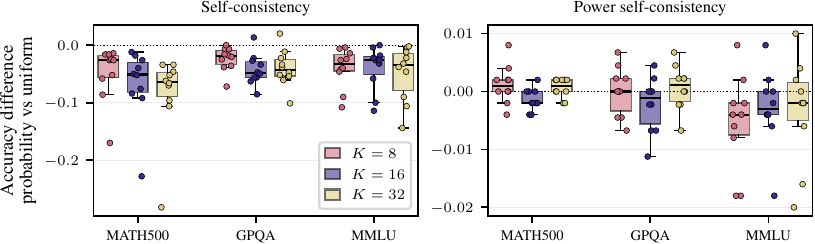}
    \caption{
    \textbf{Self-consistency with probability-weighted voting often underperforms majority voting. Power self-consistency does not shows consistent differences.}
    Accuracy difference between (power) self-consistency using uniform or probability-weighted majority voting.
    Results on \qwenthree{} model series across three datasets.
    See \Cref{sec:sub:within-sample-correlation} for details.
    }
    \label{fig:self-consistency-difference-probability-vs-uniform}
\end{figure}

\section{Discussion and Conclusion}
\label{sec:discussion}

Our results show that the relationship between sequence probability and correctness depends on the level it is measured at.
Within a fixed dataset, model, and decoding setup, higher-probability prompt-answer pairs are often more likely to be correct, suggesting that sequence probability contains useful information across examples. 
However, this signal does not generally transfer to the method-level.
When comparing responses to the same prompt, tuning hyperparameters within a method, or comparing different decoding methods, higher sequence probability does not consistently imply higher correctness. 
This is to some degree a negative result as it does not provide guidance on what method to choose or which hyperparameter to choose within a method.

Our analysis has practical implications for probability-based decoding, self-consistency, and probability-based verifier-free self-improvement.
Specifically, sequence probabilities are a useful signal, within a given dataset and across samples for a single prompt, when the model already has sufficient task accuracy.
However, it is a less useful criterion for selecting hyperparameters of methods and methods themselves.
Therefore, our findings suggest that in addition to asking \textit{whether} sequence probability correlates with correctness, it is important to ask \textit{at which granularity} it does so.

\section*{Acknowledgements}
Johannes Zenn thanks Tim Z.\ Xiao for initial discussions on log-probability and correctness.
Johannes Zenn thanks Xiaotong Ji, Rasul Tutunov, Matthieu Zimmer, and Haitham Bou Ammar for discussions on Scalable Power Sampling.
Johannes Zenn acknowledges funding by the Deutsche Forschungsgemeinschaft (DFG, German Research Foundation) under Germany’s Excellence Strategy – EXC number 2064/1 – Project number 390727645.
The authors thank the International Max Planck Research School for Intelligent Systems (IMPRS-IS) for supporting Johannes Zenn.

\bibliographystyle{plainnat}
\bibliography{references}

\clearpage
\appendix

\crefalias{section}{appendix}
\Crefname{appendix}{Appendix}{Appendices}
\crefname{appendix}{appendix}{appendices}

\crefalias{subsection}{appendix}

\numberwithin{equation}{section}
\renewcommand{\theequation}{\thesection\arabic{equation}}

\section{Related Work} 
\label{app:sec:related-work}

In this section, we give a broad overview over the related work.

\paragraph{Decoding methods.}
Search-based decoding methods methods like beam search \citep{lowerre1976harpy,sutskever2014sequence} searches over the space of sequences but often suffers from repetitions \citep{vijayakumar2016diverse,cohen_empirical_nodate}, and other deteriorations in quality \citep{koehn_six_2017,yang_breaking_2018}.
Its lack of diversity \citep{su_contrastive_2022} has been addressed by adding terms related to diversity \citep{vijayakumar2016diverse,vilnis2023arithmetic}. 
In \Cref{sec:sampling-methods}, we discuss various local decoding methods: top-$k$ sampling \citep{fan_hierarchical_2018}, top-$p$ sampling \citep{holtzman_curious_2020}, and $\varepsilon$-sampling \citep{hewitt_truncation_2022}.
\citet{kempton2025local} phrase common local decoding methods in a framework similar to our variational objective.
In this work, we instead consider a variational framework for global decoding methods.
While most prior work investigate how decoding relates to accuracy, this work asks the more general question how sequence probability and correctness are related.

\paragraph{Power sampling in LLMs.}
\citet{karan2025reasoning} introduce power sampling from LLMs as a verifier-free way to sharpen the model distribution over full sequences.
They propose a Markov Chain Monte Carlo algorithm, i.e., a Metropolis-Hastings \citep{metropolis1953equation,hastings1970monte} sampler operating in blocks of tokens.
\citet{ji2026scalable} take a similar block-based framework and approximate the token-level power distribution by using online rollouts from the top-$k$ tokens resulting in the scalable power sampling (SPS) algorithm.
\citet{azizi2026power} propose power-SMC, a sequential Monte Carlo (SMC) \citep{doucet2001introduction,del2006sequential,briers2010smoothing,chopin2020introduction} sampler, which moves a set of particles over $T$ steps (where $T$ denotes the length of the sequence), computes importance-weights, and resamples the particles.
In this work, we ask the more general question on how sequence \logprob{} and correctness are related. 
We investigate sampling from the power distribution as one global method producing more probable sequences (see \Cref{rem:power-logprob}) approximated by SPS and power-SMC.

\paragraph{The power distribution, sharpening, and reinforcement learning.}
Many works argue that reinforcement learning with verifiable rewards (RLVR) can be explained as doing distribution sharpening \citep{he2025rewarding,song2025outcome,yue2025limit,gai2025differential,ni2025origin}.
Based on this \citet{karan2025reasoning} make the point that power sampling can be compared to RLVR and show comparisons between RL-finetuned models and base models.
We, instead, show results for base and posttrained models and focus on the relationship between sequence probability and correctness instead of explicitly comparing models at different training stages.

\paragraph{Self-consistency.}
Self-consistency (SC) aggregates multiple samples from a LLM via majority voting on extracted answers \citep{wang_self-consistency_2023}. 
In cases where (simple) answer extraction is not possible, an LLM can be used to extract and compare answers \citep{wang_soft_2024,chen2024universal,wang2024integrate}. 
In this work, we analyze the correlation between probability and correctness, also \textit{within a sample}.
We find that, often, there is no consistent correlation for a single prompt, explaining why probability-weighting in SC often underperforms.

\paragraph{Choosing out of a pool of $N$ sequences.}
Best-of-$N$ \citep{cobbe2021training,lightman2024let} is a general method choosing the ``best'' sequence out of a set of $N$ sequences.
While sequences can be chosen by probability, various other confidence criteria can be defined \citep{kang_scalable_2025,fu_deep_2025}.
These criteria encode assumption about how correct sequences look like.
In this work, we investigate the underlying question asking how \logprob{} and correctness are related.

\paragraph{Calibration and log-probability.}
There are various works investigating calibration in LLMs.
\citet{spiess2025calibration} focus on the calibration of LLMs for code.
They find that code models are not well calibrated in general and they show how calibration can be improved.
\citet{kadavath2022language} show that LLMs are often well-calibrated on multiple-choice questions and questions with binary options. 
Then, they evaluate self-evaluation by asking models for answers and letting them evaluate the probability of the answer being correct.
This work is related to calibration in that it similarly computes the \logprob{} for sequences and compares it to its accuracy (within-dataset correlation).
However, we do not relate the specific sequence-probabilities to correctness and consider various other forms of correlation.
\citet{li2026leaf} motivate their tree-search method by showing that \textit{coverage} i.e., the \logprob{} of all samples generated for a given prompt, correlates with correctness.
Further, \citet{li2025sample} find that \textit{low probability tokens} correlate with large \textit{epistemic uncertainty} and propose to resort to greedily sample those tokens (as opposed to sampling from the distribution).

\clearpage
\section{Theoretical Arguments} \label{app:sec:additional-details-methods}

In this section, we provide an extended formal discussion on decoding methods highlighting various connections made in the main text.

First, we can state that beam search (without length normalization) becomes exact with large $K$.
\vskip-0.75em
\begin{repeatedremark}{rem:beam-exact}[Beam search becomes exact for $K\to\infty$]
If $K$ is chosen such that no continuation is ever pruned before a fixed $T$ then beam search returns an element of $\argmax_{\cont}\ \basemodel(\cont\mid\prompt)$.
\end{repeatedremark}
\vspace{-1.5em}
\begin{proof}
For a fixed horizon $T$, beam search keeps the $K$ highest-scoring partial continuations at each depth.
If $K$ is large enough that such that no continuation of length at most $T$ is pruned, then beam search explores the entire tree up to depth $T$.
Hence, at depth $T$, it compares all complete continuations $\cont$ and returns one with maximal probability under $\basemodel(\cont\mid\prompt)$.
\end{proof}

Power sampling, another global sampling method, increases the \textit{expected} sequence \logprob{}.
\vskip-0.75em
\begin{repeatedremark}{rem:power-logprob}[Power sampling increases expected sequence log-probability]
For $\alpha>1$, let $\cont\sim \basemodel_\alpha(\cdot\mid\prompt)$.
Then, increasing $\alpha$ shifts mass toward sequences with larger probability under $\basemodel$ as
$
\mathbb E_{\basemodel_\alpha}\left[\log \basemodel(\cont\mid\prompt)\right]
\ge
\mathbb E_{\basemodel}\left[\log \basemodel(\cont\mid\prompt)\right].
$
\end{repeatedremark}
\vspace{-1.5em}
\begin{proof}
Let
\[
Z_\alpha(\prompt)=\sum_{\cont}\basemodel(\cont\mid\prompt)^\alpha.
\]
Then, under reasonable assumptions,
\[
\frac{d}{d\alpha}\log Z_\alpha(\prompt)
=
\frac{\sum_{\cont}\basemodel(\cont\mid\prompt)^\alpha \log \basemodel(\cont\mid\prompt)}
     {Z_\alpha(\prompt)}
=
\mathbb E_{\basemodel_\alpha(\cont\mid\prompt)}\!\left[\log \basemodel(\cont\mid\prompt)\right].
\]
Differentiating once more gives
\[
\frac{d}{d\alpha}
\mathbb E_{\basemodel_\alpha(\cont\mid\prompt)}\left[\log \basemodel(\cont\mid\prompt)\right]
=
\Var_{\basemodel_\alpha(\cont\mid\prompt)}\left(\log \basemodel(\cont\mid\prompt)\right)
\ge 0.
\]
Hence $\mathbb E_{\basemodel_\alpha(\cont\mid\prompt)}[\log \basemodel(\cont\mid\prompt)]$ is nondecreasing in $\alpha$.
\end{proof}

In contrast, for \textit{local} sampling methods, we cannot guarantee any \textit{global} properties.
\vskip-0.75em
\begin{remark}[Local sampling methods do not target most probable global sequences.]
\label{rem:local-not-global}
Methods with local normalization constants modify $\basemodel(\cdot\mid \prompt,\cont_{<t})$ separately at each prefix. 
Hence, they favor continuations whose \emph{next token} is locally likely. 
This is different from maximizing the \emph{sequence-level} probability $\basemodel(\cont\mid\prompt)$.
\end{remark}
\vspace{-1.5em}
\begin{proof}
For LTS,
\begin{align}
    \nexttoken_{\alpha}(\cont\mid\prompt)
    =
    \prod_{t=1}^T \frac{\basemodel(s_t\mid \prompt,\cont_{<t})^\alpha}{\tilde Z_\alpha(\prompt,\cont_{<t})}
    =
    \basemodel(\cont\mid\prompt)^\alpha
    \prod_{t=1}^T \tilde Z_\alpha(\prompt,\cont_{<t})^{-1}.
    \label{eq:rem-local-not-global-lts}
\end{align}
For TS,
\begin{align}
    p_A(\cont\mid\prompt)
    =
    \prod_{t=1}^T
    \frac{\basemodel(s_t\mid \prompt,\cont_{<t})}{\tilde Z_A(\prompt,\cont_{<t})}
    \mathbb{1}[s_t\in A(\prompt,\cont_{<t})]
    =
    \basemodel(\cont\mid\prompt)
    \prod_{t=1}^T
    \frac{\mathbb{1}[s_t\in A(\prompt,\cont_{<t})]}{\tilde Z_A(\prompt,\cont_{<t})}.
    \label{eq:rem-local-not-global-ts}
\end{align}
In both, \Cref{eq:rem-local-not-global-lts} and \Cref{eq:rem-local-not-global-ts}, the sequence weight contains factors beyond $\basemodel(\cont\mid\prompt)$ itself.
This is different from the normalization constant $Z_\alpha(\bar\cont) = \sum_\rvs p(\rvs\mid\bar\rvs)^\alpha$ (\Cref{eq:power-dist} in \Cref{sec:sampling-methods}).
\end{proof}

LTS defines a temperature $\alpha$ similar to power-sampling.
However, there is, in general, no temperature for LTS that recovers samples from the power distribution.
\vskip-0.75em
\begin{repeatedremark}{rem:lts-not-power}[LTS is generally not equal to power sampling.]
Unless the local normalization constants of LTS are identical across all reachable prefixes, no single token-level temperature reproduces the sequence-level power distribution.
\end{repeatedremark}
\vspace{-1.5em}
\begin{proof}
The LTS-induced sequence distribution is
\begin{align}
    \nexttoken_\alpha(\cont\mid\prompt)
    =
    \prod_{t=1}^T \frac{\basemodel(s_t\mid \prompt,\cont_{<t})^\alpha}
    {\tilde Z_\alpha(\prompt,\cont_{<t})}
    =
    \frac{\basemodel(\cont\mid\prompt)^\alpha}
    {\prod_{t=1}^T \tilde Z_\alpha(\prompt,\cont_{<t})}.
\end{align}
By contrast, the sequence-level power distribution is
\[
\basemodel_\alpha(\cont\mid\prompt)
=
\frac{\basemodel(\cont\mid\prompt)^\alpha}{Z_\alpha(\prompt)}.
\]
Thus LTS coincides with sequence-level power sampling only if
\[
\prod_{t=1}^T \tilde Z_\alpha(\prompt,\cont_{<t})
\overset{!}{=}
Z_\alpha(\prompt) 
\;
\forall \; (\prompt, \cont_{<t}).
\]
However, the local normalization constants depend on the prefix, so the product is path-dependent and LTS is not equal to sequence-level power sampling.
\end{proof}

As discussed in the main test, we find the power distribution to be the maximizer of $\mathcal J_\alpha(q)$.
\vskip-0.75em
\begin{repeatedremark}{rem:power-distribution-minimizer}[The power distribution is a maximizer of the variational objective.]
The maximizer of $\mathcal J_\alpha(q)$ is the power distribution,
$
    q_\alpha^\star
    \in
    \argmax_q \mathcal{J}_\alpha(q)
    \Longleftrightarrow
    q_\alpha^\star(\cont\mid\prompt)
    =
    \basemodel_\alpha(\cont\mid\prompt).
$
\end{repeatedremark}
\vspace{-1.5em}
\begin{proof}
We have
$
    \mathcal J_\alpha(q)
    =
    \sum_{\cont} q(\cont\mid\prompt)\log \basemodel(\cont\mid\prompt)
    -
    \frac{1}{\alpha}
    \sum_{\cont} q(\cont\mid\prompt)\log q(\cont\mid\prompt),
$
and,
\begin{align}
    \mathcal J_\alpha(q)
    &=
    -\frac{1}{\alpha}
    \sum_{\cont} q(\cont\mid\prompt)
    \log
    \frac{q(\cont\mid\prompt)}{\basemodel(\cont\mid\prompt)^\alpha}
    \\
    &=
    -\frac{1}{\alpha}
    \sum_{\cont} q(\cont\mid\prompt)
    \log
    \frac{q(\cont\mid\prompt)}{\basemodel_\alpha(\cont\mid\prompt) Z_\alpha(\prompt)}
    \\
    &=
    -\frac{1}{\alpha}
    \KL\!\bigl(q(\cdot\mid\prompt)\,\|\,\basemodel_\alpha(\cdot\mid\prompt)\bigr)
    +
    \frac{1}{\alpha}\log Z_\alpha(\prompt),
\end{align}
where
\[
    Z_\alpha(\prompt)
    =
    \sum_{\cont}\basemodel(\cont\mid\prompt)^\alpha.
\]
Since $\log Z_\alpha(\prompt)$ does not depend on $q$, maximizing $\mathcal J_\alpha(q)$ is equivalent to minimizing
\[
    \KL\!\bigl(q(\cdot\mid\prompt)\,\|\,\basemodel_\alpha(\cdot\mid\prompt)\bigr),
\]
whose unique minimizer is $q(\cdot\mid\prompt)=\basemodel_\alpha(\cdot\mid\prompt)$.
\end{proof}

Investigating the hyperparameter $\alpha$, we find a tradeoff between expected \logprob{} and entropy.
\vskip-0.75em
\begin{remark}[Expected log-probability increases and entropy decreases with $\alpha$]
For $\alpha>1$, increasing $\alpha$ shifts mass toward higher-probability continuations under $\basemodel$.
Therefore,
$
\mathbb E_{\basemodel_\alpha}\!\left[\log \basemodel(\cont\mid\prompt)\right]
$
is monotonically increasing in $\alpha$, while
$
\mathcal H(\basemodel_\alpha)
$
is monotonically decreasing in $\alpha$.
\end{remark}
\vspace{-1.5em}
\begin{proof}
We have seen the first part already in \Cref{rem:power-logprob}.
For the entropy, we obtain
\begin{align}
    \mathcal H(\basemodel_\alpha)
    &=
    -\sum_{\cont}
    \basemodel_\alpha(\cont\mid\prompt)
    \log \basemodel_\alpha(\cont\mid\prompt)
    \\
    &=
    -\sum_{\cont}
    \basemodel_\alpha(\cont\mid\prompt)
    \Bigl(
        \alpha \log \basemodel(\cont\mid\prompt)
        -
        \log Z_\alpha(\prompt)
    \Bigr)
    \\
    &=
    -\alpha\,
    \mathbb E_{\basemodel_\alpha}
    \!\left[
        \log \basemodel(\cont\mid\prompt)
    \right]
    +
    \log Z_\alpha(\prompt).
\end{align}
Differentiating gives
\begin{align}
    \frac{\partial}{\partial \alpha}\mathcal H(\basemodel_\alpha)
    &=
    -\alpha\,
    \frac{\partial}{\partial \alpha}
    \mathbb E_{\basemodel_\alpha}
    \!\left[
        \log \basemodel(\cont\mid\prompt)
    \right]
    \\
    &=
    -\alpha\,
    \Var_{\basemodel_\alpha}
    \!\left(
        \log \basemodel(\cont\mid\prompt)
    \right)
    \le 0.
\end{align}
Therefore, $\mathcal H(\basemodel_\alpha)$ is monotonically decreasing in $\alpha$.
\end{proof}

\clearpage
\section{A Twisted Sequential Monte Carlo Framework for Power Sampling}
\label{app:sec:twisted-smc-framework}

Sequential Monte Carlo (SMC) algorithms \citep{doucet2001introduction,del2006sequential,briers2010smoothing,chopin2020introduction} are one of the main algorithms for estimating (global) normalization constants.
In this section, we show how recent work \citep{ji2026scalable,azizi2026power} fits into the \textit{twisted SMC framework} of \citet{zhao2024probabilistic}.

Let
$
\gamma(\cont \mid \prompt) \coloneqq \basemodel(\cont \mid \prompt)^\alpha
$
denote the target distribution. 
Since exact sampling from the normalized distribution induced by $\gamma$ is intractable, a natural approach is (twisted) SMC which constructs a sequence intermediate target distributions and propagates a population of particles.
Twisted SMC \citep{zhao2024probabilistic} introduces positive twist functions $\psi_t(\prompt,\cont_{\le t})$ and defines
$
    \gamma_t(\cont_{\le t}\mid \prompt)
    \coloneqq
    \basemodel(\cont_{\le t}\mid \prompt)^\alpha \,\psi_t(\prompt,\cont_{\le t})
$
for $t<T$, with $\gamma_T=\gamma$. 
For a proposal $q_t(s_t\mid \prompt,\cont_{<t})$, the incremental weight is
$
    w_t^{i}
    \propto
    w_{t-1}^{i}
    {\basemodel(s_t^{i}\mid \prompt,\cont_{<t}^{i})^\alpha}/
         {q_t(s_t^{i}\mid \prompt,\cont_{<t}^{i})}
    {\psi_t(\prompt,\cont_{\le t}^{i})}/
         {\psi_{t-1}(\prompt,\cont_{<t}^{i})}.
$
\begin{remark}[Optimal twist following {\citet[Proposition 3.2]{zhao2024probabilistic}}] 
The optimal twist functions targeting the power distribution are given as
\begin{align}
    \psi_t^\star(\prompt,\cont_{\le t})
    \propto
    \sum_{\cont_{>t}}
    \basemodel(\cont_{>t}\mid \prompt,\cont_{\le t})^\alpha.
    \label{eq:twisted-smc-optimal-value}
\end{align}
\end{remark}
\vspace{-1.5em}
\begin{proof}
The terminal unnormalized target is
\[
\gamma_T(\cont\mid\prompt)=\basemodel(\cont\mid\prompt)^\alpha.
\]
For a fixed prefix $\cont_{\le t}$, summing out the future suffix gives
\begin{align}
    \sum_{\cont_{>t}} \gamma_T(\cont\mid\prompt)
    &=
    \sum_{\cont_{>t}}
    \Bigl(
        \basemodel(\cont_{\le t}\mid\prompt)^\alpha
        \basemodel(\cont_{>t}\mid\prompt,\cont_{\le t})^\alpha
    \Bigr) \\
    &=
    \basemodel(\cont_{\le t}\mid\prompt)^\alpha
    \sum_{\cont_{>t}}
    \basemodel(\cont_{>t}\mid\prompt,\cont_{\le t})^\alpha.
\end{align}
Therefore, choosing
\[
\psi_t^\star(\prompt,\cont_{\le t})
\propto
\sum_{\cont_{>t}}
\basemodel(\cont_{>t}\mid\prompt,\cont_{\le t})^\alpha
\]
makes
\[
\gamma_t(\cont_{\le t}\mid\prompt)
=
\basemodel(\cont_{\le t}\mid\prompt)^\alpha\psi_t^\star(\prompt,\cont_{\le t})
\]
proportional to the exact marginal of $\gamma_T$ at time $t$.
\end{proof}

The two recently proposed methods fall into this framework as we explain next.

\paragraph{Power-SMC.}
Power-SMC \citep{azizi2026power} corresponds to the \emph{untwisted} choice $\psi_t\equiv 1$ using the LTS proposal distribution
$
    q_t(s_t\mid \prompt,\cont_{<t})
    =\nexttoken_\alpha(s_t\mid\prompt,\cont_{<t})
$.
While this minimizes the variance among prefix-only proposals \citep{azizi2026power}, it does not account for future continuation mass.

\begin{remark}[Asymptotic exactness of Power-SMC following \citet{chopin2004central} and \citet{delmoral2004feynmankac}.]
For any fixed horizon, consistency of SMC implies that, under the usual regularity assumptions, the empirical particle approximation converges to the target distribution as $N\to\infty$.
\end{remark}
\vspace{-1.5em}
\begin{proof}
Power-SMC is a standard SMC algorithm for the Feynman-Kac model with unnormalized targets
\[
\gamma_t(\cont_{\le t}\mid\prompt)
=
\basemodel(\cont_{\le t}\mid\prompt)^\alpha
\]
and proposal
\[
q_t(s_t\mid\prompt,\cont_{<t})=\nexttoken_\alpha(s_t\mid\prompt,\cont_{<t}).
\]
Therefore the usual SMC law of large numbers applies.
We refer to \citet{chopin2004central} and \citet{delmoral2004feynmankac} for the corresponding theoretical arguments.
\end{proof}

\paragraph{Scalable power sampling (SPS).}
Similarly, SPS \citep{ji2026scalable} corresponds to a \emph{twisted} SMC method that approximates the optimal term $\psi_t^\star(\prompt,\cont_{<t},s_t)$ in \Cref{eq:twisted-smc-optimal-value} via Monte Carlo sampling restricted to a top-$k$ candidate set.
SPS utilizes a blockwise approximation for a block $b_t=\cont_{(t-1)B+1:tB}$ of $B$ tokens,
$
    \basemodel_\alpha^{\psi_\star}(b_t\mid \prompt,\cont_{1:(t-1)B})
    \propto
    \basemodel(b_t\mid \prompt,\cont_{1:(t-1)B})^\alpha\,
    \psi_{tB}^\star(\prompt,\cont_{1:(t-1)B}\oplus b_t),
$.

\begin{remark}[Approximation gaps for SPS]
SPS is exact if the look-ahead term $\psi^\star$ is evaluated exactly at every position.
Alternatively, it is exact when the candidate set is not truncated and the next block is sampled exactly from the corresponding blockwise fully-adapted conditional.
In practice, the approximation comes from (i) Monte Carlo estimation of $\psi^\star$, (ii) top-$k$ truncation, and (iii) the blockwise restriction.
\end{remark}
\vspace{-1.5em}
\begin{proof} 
For a block $b_t$, the exact fully adapted block proposal is proportional to
\[
\basemodel(b_t\mid \prompt,\cont_{1:(t-1)B})^\alpha\,
\psi_{tB}^\star(\prompt,\cont_{1:(t-1)B}\oplus b_t).
\]
If $\psi^\star$ is evaluated exactly and sampling is performed exactly from this conditional over the full candidate set, then the corresponding twisted SMC update is exact.
Likewise, if no top-$k$ truncation is applied and the next block is sampled exactly from the blockwise fully adapted conditional, then no approximation is introduced at that block update.

SPS departs from this exact construction in three places:
(i) it replaces $\psi^\star$ by a Monte Carlo estimate,
(ii) it truncates the candidate set to a top-$k$ subset, and
(iii) it uses a blockwise approximation rather than the exact full-sequence update.
\end{proof} 

The twisted SMC view of \citet{zhao2024probabilistic} shows interesting directions for future work.
Improvements can be mostly twofold:
Firstly, one can \textit{learn the twist functions} during inference via various methods proposed in the SMC literature \citep{zhao2024probabilistic,lawson2022sixo,lioutascritic}.
Orthogonally, one can extend the framework with verifiers or reward models.

\clearpage
\section{Details on Experimental Setup}
\label{app:sec:experimental-setup}

In this section we discuss our implementation of sampling methods in \Cref{app:sec:implementation}.
Then we discuss the correlation coefficients in \Cref{app:sec:sub:correlation-coefficients}, canonical hyperparameters in \Cref{app:sec:sub:canonical-hyperparameters}, and our computational requirements and corresponding datasets in \Cref{app:sec:sub:computational-requirements-datasets}.
We provide prompt templates in \Cref{app:sec:prompt-template}.

\subsection{Implementation}
\label{app:sec:implementation}

We use most of the local sampling methods as implemented in the vLLM \citep{kwon2023efficient} framework.
For power-SMC we use the code as provided by the authors \citep{azizi2026power}.
We implement SPS following the algorithm provided in the paper \citep{ji2026scalable}.
We do rollouts with LTS and $\alpha=4.0$ and correct this bias by using importance weighting to estimate the $\zeta$ functions.
Further, we set $K=M=8$ and $\tau_c=\tau_r=0.25$ as per the original work \citep{ji2026scalable}.
\Cref{alg:importance-weighted-sps} makes the exact algorithm concrete.

\begin{algorithm}[h]
  \caption{Importance-weighted SPS}
  \label{alg:importance-weighted-sps}
  \begin{algorithmic}[1]
    \Require prompt $\prompt$; base model $p$; candidate temperature\ $\tau_c$; rollout temperature\ $\tau_r$;
      block size $B$; top $K$; candidates per block $M$; power exponent $\alpha$; rollouts $R$; rollout horizon $H$;
      max tokens $T$
    \State $\cont \gets \emptyset$;\ $t \gets 0$
    \While{$t < T$ \textbf{and} not finished}
      \State $B_t \gets \min(B,\ T-t)$
      \Statex \textit{\quad // 1. Sample with p and temperature $\tau_c$, score under $p$}
      \State Draw $\{\tilde\cont_i\}_{i=1}^{M} \sim q(\cdot \mid \prompt, \cont)$ of $\leq B_t$ tokens where $q$ uses $p$ to sample with temp. $\tau_c$
      \State Sort $\{\tilde\cont_i\}$ by probability under $p$ and keep top $K$
      \Statex \textit{\quad // 2. Estimate $\zeta$}
      \For{$i = 1, \dots, K$}
        \State Draw $\{\rvr_{i,j}\}_{j=1}^R \sim q(\cdot \mid \prompt, \cont, \tilde\cont_i)$ of $\leq H$ tokens where $q$ uses $p$ to sample with temp. $\tau_r$
        \State $\log u_{i,j} \gets \alpha\log p(\rvr_{i,j} \mid \prompt, \cont, \tilde\cont_i) - \log q(\rvr_{i,j} \mid \prompt, \cont, \tilde\cont_i)$
        \State $\log\hat\zeta_i \gets \log\Big(\tfrac{1}{R}\sum_j \exp(\log u_{i,j})\Big)$
        \For{$s = 1, \dots, R$}
          \State $\log\hat\zeta_i^{(-s)} \gets \log\!\Big(\tfrac{1}{R-1}\sum_{j\neq s}\exp(\log u_{i,j})\Big)$
        \EndFor
      \EndFor
      \State Compute the Jackknife bias correction using $\log\hat\zeta_i$ and $\log\hat\zeta_i^{(-s)}$
      \Statex \textit{\quad // 3. Select block}
      \State $\log b_i \gets \alpha\log p(\tilde\cont_i \mid \prompt, \cont)$
      \State $\pi_i \propto \exp(\log b_i + \log\hat\zeta_i)$
      \State Sample $i^\star \sim \pi$;\ $\rvs \gets \rvs | \tilde\rvs_i$;\ $t \gets t + | \tilde\rvs_{i^\star}|$
      \State Mark finished if an end-of-sequence token appears in  stop token was emitted in $ \tilde\rvs_{i^\star}$
    \EndWhile
    \State \Return $\rvs$
  \end{algorithmic}
  \end{algorithm}

\subsection{Correlation Coefficients}
\label{app:sec:sub:correlation-coefficients}

We compute correlations between the continuous \logprob{} and a binary correctness label $y\in\{0, 1\}$.
The per-sample Spearman rank correlation $\rho$ \citep{spearman1904proof} measures monotone agreement between the two quantities.
Since $y$ is binary, the $\rho$ is upper bound by $\sqrt{0.5}$.
The binned Pearson correlation coefficient \citep{pearson1894philosophical} $r$ partitions \logprob{} into $K=10$ equal-count bins, computes the bin-mean (for \logprob{} and correctness) and computes the ordinary Pearson coefficient $r$.

\subsection{Canonical Hyperparameters}
\label{app:sec:sub:canonical-hyperparameters}

\Cref{tab:canonical-hyperparameters} lists the canonical hyperparameters that we use in the main text.

\begin{table}[h]
    \centering
    \begin{tabularx}{\linewidth}{L L}
        \toprule
        Decoding method & Hyperparameter \\
        \midrule
        LTS & $\alpha=4$\\
        Bo$N$ & $N=32$\\
        SPS & $B=192$\\
        Power-SMC & $N=32$\\
        Top-$k$ & $k=8$\\
        Top-$p$ & $p=0.9$\\
        Beam search & $N=2$\\
        $\varepsilon$-sampling & $\varepsilon=0.0009$\\
        \bottomrule
    \end{tabularx}
    \caption{Canonical hyperparameters for each method.}
    \label{tab:canonical-hyperparameters}
\end{table}

\subsection{Computational Requirements and Datasets}
\label{app:sec:sub:computational-requirements-datasets}

We set our computational budget to $\leq 1$ days per method and dataset using an NVIDIA H$100$ graphics card with $80$G of memory.
To fulfill this budget, we use $\leq 500$ samples of each dataset.
For MMLU \citep{wang_mmlu-pro_2024}, this corresponds to the splits ``abstract algebra'', ``anatomy'', and ``astronomy''.
Further, we evaluate all datasets and models on a \textit{shared set of prompts} that they generate a non-empty response for.
All of these restrictions result in the numbers reported in the numbers of samples per dataset as reported in \Cref{tab:number-prompts-per-dataset}.
Empty cells in our summary plots are runs that did not finish within the compute time limits.

\begin{table}[h]
    \centering
    \begin{tabularx}{\linewidth}{L L}
        \toprule
        Dataset & Number of prompts \\
        \midrule
        GPQA      & $445$ \\
        HumanEval & $500$ \\
        IFEval    & $500$ \\
        MATH500   & $500$ \\
        MedQA     & $500$ \\
        MMLU      & $500$ \\
        \bottomrule
    \end{tabularx}
    \caption{Number of prompts per dataset for all plots.}
    \label{tab:number-prompts-per-dataset}
\end{table}

\clearpage
\subsection{Prompt Templates}
\label{app:sec:prompt-template}

We use the following prompt templates for each benchmark.
We prompt base models directly with the prompt and use the chat template for posttrained models.

\textbf{Math500.}

\begin{quote}
\ttfamily
Can you solve the following math problem? \{problem\}\\
Please reason step by step, and put your final answer within
\textbackslash boxed\{\}.
\end{quote}

\textbf{Humaneval.}

\begin{quote}
\ttfamily
Write a Python function to solve the following problem:

\{prompt\}
\end{quote}

\textbf{GPQA.}

\begin{quote}
\ttfamily
Answer the following multiple choice question. The last line of
your response should be of the following format:
'\textbackslash boxed\{LETTER\}' (without quotes) where LETTER is one of
ABCD (e.g.\ '\textbackslash boxed\{A\}'). Think step by step before answering.

\{question\}

A) \{choice\_a\}\\
B) \{choice\_b\}\\
C) \{choice\_c\}\\
D) \{choice\_d\}
\end{quote}

\textbf{MMLU.}

\begin{quote}
\ttfamily
Answer the following multiple-choice question. The last line of
your response should be of the following format:
'\textbackslash boxed\{LETTER\}' (without quotes) where LETTER is one of
ABCD (e.g.\ '\textbackslash boxed\{A\}'). Think step by step before answering.

\{question\}

A) \{choice\_a\}\\
B) \{choice\_b\}\\
C) \{choice\_c\}\\
D) \{choice\_d\}
\end{quote}

\textbf{MedQA.}

\begin{quote}
\ttfamily
Answer the following medical multiple-choice question. The last line of
your response should be of the following format:
'\textbackslash boxed\{LETTER\}' (without quotes) where LETTER is one of
ABCD (e.g.\ '\textbackslash boxed\{A\}'). Think step by step before answering.

\{question\}

A) \{choice\_a\}\\
B) \{choice\_b\}\\
C) \{choice\_c\}\\
D) \{choice\_d\}
\end{quote}

\textbf{IFEval.}

\begin{quote}
\ttfamily
\{prompt\}
\end{quote}

\clearpage
\section{Additional Experiments and Analysis}
\label{app:sec:additional-experiments-analysis}

In this section we provide additional experiments and analysis.
\Cref{app:sec:sub:correlation-coefficient} shows correlation coefficients for within-dataset correlation using Spearman $\rho$ instead of Pearson $r$.
\Cref{app:sec:other-models} shows results for the model families (\qwentwofive{}, \olmo{}) not listed in the main text.
\Cref{app:sec:within-sample-correlation} shows further and more granular results for within-sample correlation and discusses within-sample correlation results across methods.
\Cref{app:sec:sequence-level} discusses \logprobs{} globally, on a sequence-level, and compares \logprob{} to sequence-normalized \logprob{}.
\Cref{app:sec:sub:sequence-lengths} discusses sequence lengths across methods.
\Cref{sec:sub:mode-vs-power-distribution} discusses the experimental results between mode and power distribution.
\Cref{app:sec:sub:self-distillation} shows results for self-distillation.
\Cref{app:sec:sub:qwen3-thinking} shows results for the \qwenthree{} thinking models.

\subsection{Computing Correlation}
\label{app:sec:sub:correlation-coefficient}

In this section, we compare the correlation coefficients that we use in the main text for measuring within-dataset correlation (see \Cref{sec:sub:within-sample-correlation}).

When comparing the per-sample Spearman correlation coefficient $\rho$ in \Cref{fig:family-heatmap-per-sample-spearman-aggregate} to \Cref{fig:family-heatmap-bin-pearson-aggregate}, we find that both plots show the same structure but Pearson correlation coefficients $r$ seem to be more extreme.

\begin{figure}[h]
    \centering
    \includegraphics{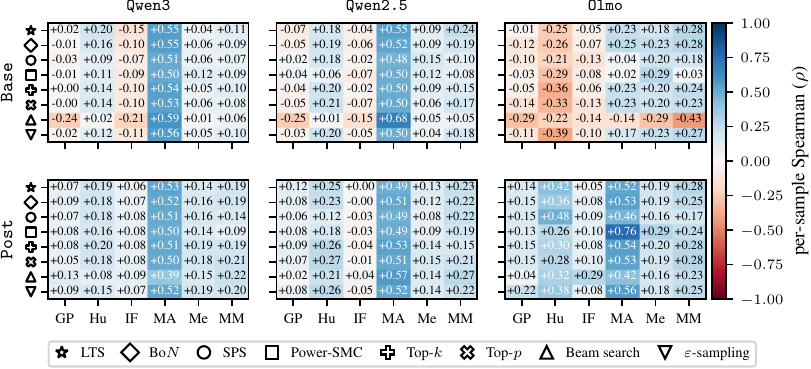}
    \caption{\textbf{We observe consistent within-dataset correlation across model families and datasets largely independent of methods.} 
    \texttt{Base} models show more negative and diverse correlation and are consistently negative for IFEval.
    \texttt{Posttrained} models show consistently positive correlation.
    Correlation coefficients $\rho$ averaged over model sizes at representative hyperparameter for each method.
    Methods are in the legend.
    Datasets are plotted horizontally on the bottom of each panel:
    \underline{GP}QA,
    \underline{Hu}maneval,
    \underline{IF}Eval,
    \underline{MA}TH500,
    \underline{Me}dQA,
    \underline{MM}LU.
    }
    \label{fig:family-heatmap-per-sample-spearman-aggregate}
\end{figure}

\clearpage
\subsection{Other Models}
\label{app:sec:other-models}

We provide the missing models for within-method correlation (\Cref{app:sec:sub:correlation-method-model-dataset-within-hyperparameter}) and across method correlation (\Cref{app:sec:sub:correlation-across-methods}).

\subsubsection{Within-Method Correlation}
\label{app:sec:sub:correlation-method-model-dataset-within-hyperparameter}

\Cref{fig:correlation-within-method-qwen25} and \Cref{fig:correlation-within-method-olmo} show the within-method correlation results for \qwentwofive{} and \olmo{}.

\begin{figure}[h]
    \includegraphics{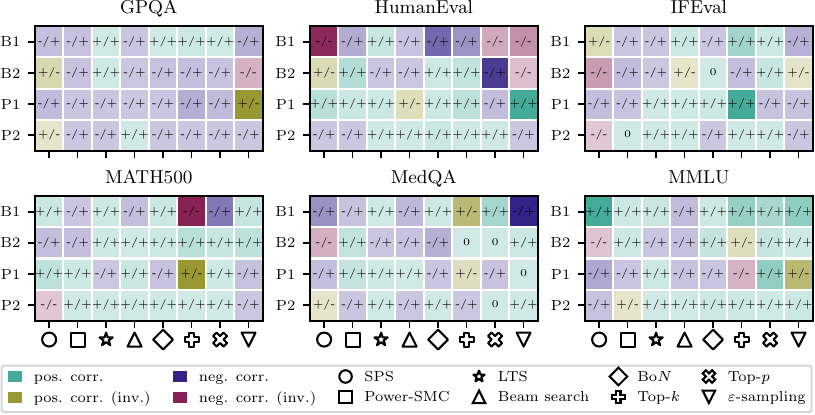}
    \caption{
    \textbf{Methods show either positive or negative correlation within their hyperparameter.}
    Correlations within methods for local and global decoding methods, models (\qwentwofive{} series $8\mathrm B$ as $\ast$1, \texttt{Math} $8\mathrm B$ as $\ast$2 with base as B$\ast$ and posttrained model as P$\ast$), and benchmark datasets.
    Correlations are not consistent across datasets and models.
    Detailed discussion in \Cref{sec:correlation-within-method-across-hp}.
    }
    \label{fig:correlation-within-method-qwen25}
\end{figure}

\begin{figure}[h]
    \includegraphics{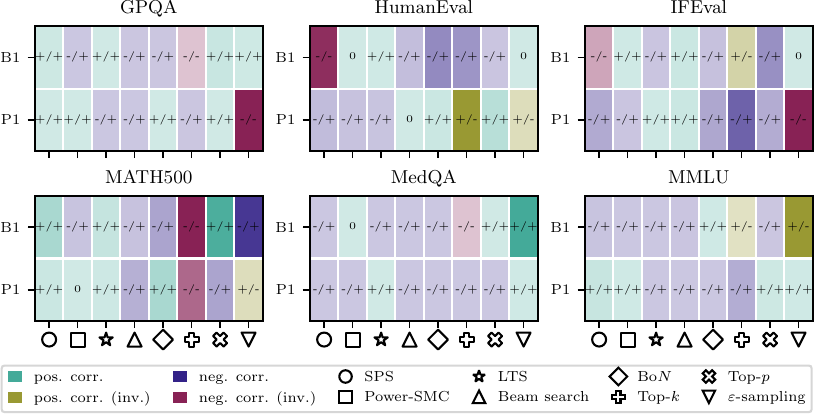}
    \caption{
    \textbf{Methods show either positive or negative correlation within their hyperparameter.}
    Correlations within methods for local and global decoding methods, models (\olmo{} series $7\mathrm B$ as $\ast$1 with base as B$\ast$ and posttrained model as P$\ast$), and benchmark datasets.
    Correlations are not consistent across datasets and models.
    Detailed discussion in \Cref{sec:correlation-within-method-across-hp}.
    }
    \label{fig:correlation-within-method-olmo}
\end{figure}

\subsubsection{Across-Method Correlation}
\label{app:sec:sub:correlation-across-methods}

\Cref{fig:correlation-accuracy-by-method-dataset-qwen25} and \Cref{fig:correlation-accuracy-by-method-dataset-olmo} show the across-method correlation results for \qwentwofive{} and \olmo{}.

\begin{figure}[h]
    \includegraphics{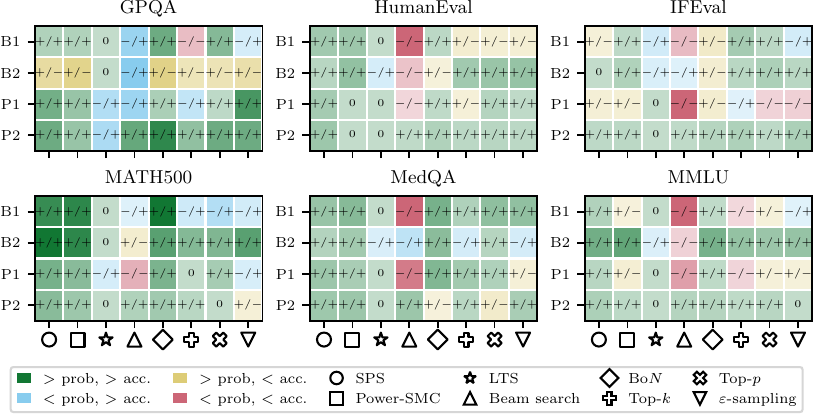}
    \caption{
    \textbf{Various datasets and methods show positive correlations between sequence probability and correctness.}
    Correlations across methods for local and global decoding methods, models (\qwentwofive{} series $8\mathrm B$ as $\ast$1, \texttt{Math} $8\mathrm B$ as $\ast$2 with base as B$\ast$ and posttrained model as P$\ast$), and benchmark datasets.
    Correlations are not consistent across datasets and models.
    See \Cref{sec:sub:correlation-across-methods}.
    }
    \label{fig:correlation-accuracy-by-method-dataset-qwen25}
\end{figure}

\begin{figure}[h]
    \includegraphics{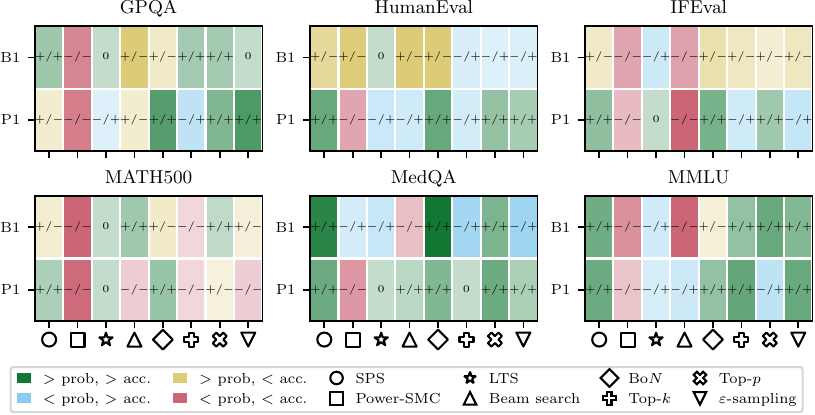}
    \caption{
    \textbf{Various datasets and methods show positive correlations between sequence probability and correctness.}
    Correlations across methods for local and global decoding methods, models(\olmo{} series $7\mathrm B$ as $\ast$1 with base as B$\ast$ and posttrained model as P$\ast$), and benchmark datasets.
    Correlations are not consistent across datasets and models.
    See \Cref{sec:sub:correlation-across-methods}.
    }
    \label{fig:correlation-accuracy-by-method-dataset-olmo}
\end{figure}

\clearpage
\subsection{Within-Sample Correlation}
\label{app:sec:within-sample-correlation}

In this section we provide additional results for the within-sample correlation investigated in \Cref{sec:sub:within-sample-correlation}.
Specifically, we investigate the results from \Cref{sec:sub:within-sample-correlation} more deeply in \Cref{app:sec:subsub:within-sample-correlation-within-method} and, further, discuss within-sample correlation across-methods in \Cref{app:sec:subsub:within-sample-correlation-across-methods}.

\subsubsection{Within-Sample Correlation Within Method}
\label{app:sec:subsub:within-sample-correlation-within-method}

\begin{wrapfigure}{r}{0.5\textwidth}
    \vskip-0.5em
    \includegraphics{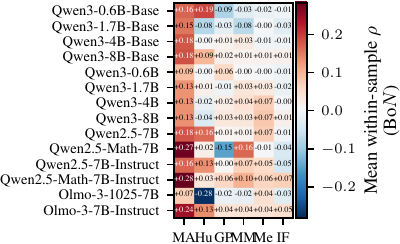}
    \caption{
        \textbf{Within-sample correlation coefficients are generally small with MATH500 consistently showing much larger values.}
         Datasets are plotted horizontally on the bottom:
        \underline{GP}QA,
        \underline{Hu}maneval,
        \underline{IF}Eval,
        \underline{MA}TH500,
        \underline{Me}dQA,
        \underline{MM}LU.
    }
    \label{fig:within-sample-corelation-coefficients-per-model-dataset}
\end{wrapfigure}
In this section, we extend our analysis of \Cref{sec:sub:within-sample-correlation}.
\Cref{fig:within-sample-corelation-coefficients-per-model-dataset} shows mean correlation coefficients of \Cref{fig:within-sample-correlation-lts-boxplot-difficulty} (left, middle).
We are able to consistently verify our results from the corresponding section. 
Specifically, we find the boxplots to summarize the granular results in \Cref{fig:within-sample-corelation-coefficients-per-model-dataset} well:
While most of the correlation coefficients are of small magnitude, only for MATH500 we consistently observe positive correlation.

Moving to \Cref{fig:within-sample-correlation-lts-boxplot-difficulty} (right), we provide in \Cref{fig:correlation-vs-fraction-correct-model-family-variant} a finer-grained analysis further splitting the data according to model family (columns) and model variants (rows).
We consistently observe a similar pattern showing that for more more correct samples, also the correlation coefficient within the samples increases on average.
Standard deviations are large and overlapping.
The order of the means over methods changes between model families.
MATH500 always seems to have highest mean.

\begin{figure}[h]
    \includegraphics{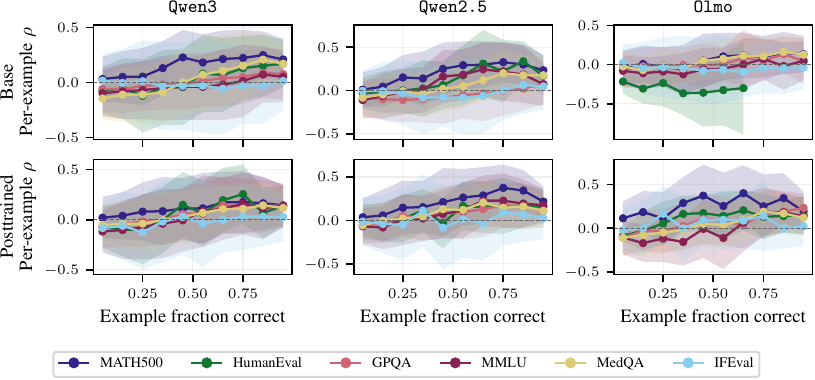}
    \caption{
    \textbf{The more correct a sample is under repeated sampling, the stronger its correlation coefficient.}
    We observe similar results across all model families (columns) and variants (rows).
    }
    \label{fig:correlation-vs-fraction-correct-model-family-variant}
\end{figure}

\clearpage
\subsubsection{Within-Sample Correlation Across Methods}
\label{app:sec:subsub:within-sample-correlation-across-methods}

Next, we investigate within-sample correlation across methods.
To do so, we compute correlation coefficients across methods, for each sample of the dataset separately.
For each method we choose a canonical hyperparameter (\Cref{tab:canonical-hyperparameters}).

\begin{figure}[h]
    \includegraphics{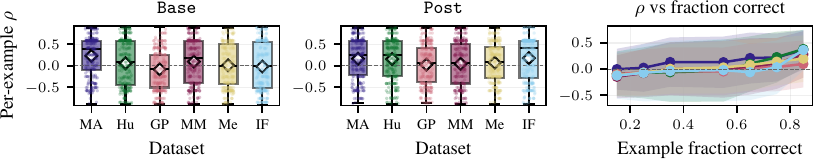}
    \caption{
    \textbf{Correlation coefficients are mostly symmetrically distributed around zero, the more correct a sample is, the more positive its correlation.}
    Per-sample rank correlation coefficient of \qwenthree{} \texttt{base} models (left) and \texttt{posttrained} models (middle) is distributed symmetrically with mean zero.
    Correlation is computed within a sample \texttt{across methods}
    Right: 
    Correlation coefficient and fraction of correct samples seems positively correlated.
    Datasets are plotted horizontally on the bottom:
        \underline{GP}QA,
        \underline{Hu}maneval,
        \underline{IF}Eval,
        \underline{MA}TH500,
        \underline{Me}dQA,
        \underline{MM}LU.
    }
    \label{fig:within-sample-correlation-across-methods-boxplot-difficulty}
\end{figure}

\begin{figure}[h]
    \includegraphics{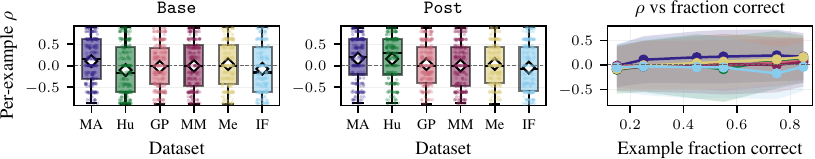}
    \caption{
    \textbf{Correlation coefficients are mostly symmetrically distributed around zero, the more correct a sample is, the more positive its correlation.}
    Per-sample rank correlation coefficient of \qwenthree{} \texttt{base} models (left) and \texttt{posttrained} models (middle) is distributed symmetrically with mean zero.
    Correlation is computed within a sample \texttt{across methods}
    Right: 
    Correlation coefficient and fraction of correct samples seems positively correlated.
    Same as \Cref{fig:within-sample-correlation-across-methods-boxplot-difficulty} but excluding Beam search.
    Datasets are plotted horizontally on the bottom:
        \underline{GP}QA,
        \underline{Hu}maneval,
        \underline{IF}Eval,
        \underline{MA}TH500,
        \underline{Me}dQA,
        \underline{MM}LU.
    }
    \label{fig:within-sample-correlation-across-methods-boxplot-difficulty-no-beam}
\end{figure}

\begin{figure}[h]
    \includegraphics{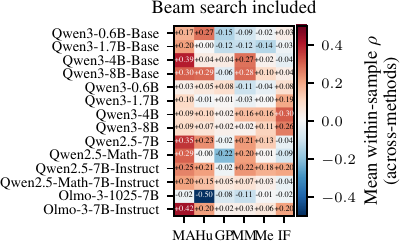}
    \includegraphics{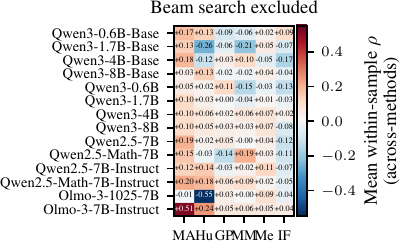}
    \caption{
        \textbf{Within-sample correlation coefficients are generally small with MATH500 consistently showing much larger values.}
         Datasets are plotted horizontally on the bottom:
        \underline{GP}QA,
        \underline{Hu}maneval,
        \underline{IF}Eval,
        \underline{MA}TH500,
        \underline{Me}dQA,
        \underline{MM}LU.
    }
    \label{fig:within-sample-corelation-coefficients-across-methods-per-model-dataset}
\end{figure}

We show our results in \Cref{fig:within-sample-correlation-across-methods-boxplot-difficulty} and \Cref{fig:within-sample-correlation-across-methods-boxplot-difficulty-no-beam} where the former include beam search and the latter do not include beam search.
We do this distinction as beam search often behaves like an outlier and might influence our conclusions.
We show the raw correlation coefficients in \Cref{fig:within-sample-corelation-coefficients-across-methods-per-model-dataset}.
From the distribution of within-sample correlation we again observe a vast range of values distributed around a zero mean.
Generally, beam search indeed seems to act like an outlier leading to jumps in the correlation coefficients (even turning signs).
Generally, the correlation coefficients seem to be similarly distributed as in the within-method case in \Cref{fig:within-sample-corelation-coefficients-per-model-dataset}.

Focusing on the fractions of correct samples plotted against the correlation coefficients, as shown in \Cref{fig:correlation-vs-fraction-correct-across-methods-model-family-variant} and \Cref{fig:correlation-vs-fraction-correct-across-methods-model-family-variant-no-beam} we find less correlation and even larger standard deviations.
When excluding beam search \Cref{fig:correlation-vs-fraction-correct-across-methods-model-family-variant-no-beam} curves seem to be mostly flat.
Given the small sample size, we should not draw any conclusions from these results.

\begin{figure}[h]
    \includegraphics{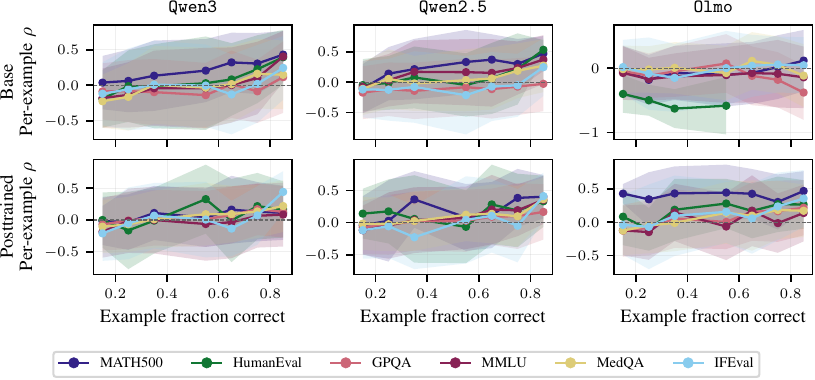}
    \caption{
    \textbf{The more correct a sample is under repeated sampling, the stronger its correlation coefficient.}
    We observe similar results across all model families (columns) and variants (rows).
    }
    \label{fig:correlation-vs-fraction-correct-across-methods-model-family-variant}
\end{figure}  

\begin{figure}[h]
    \includegraphics{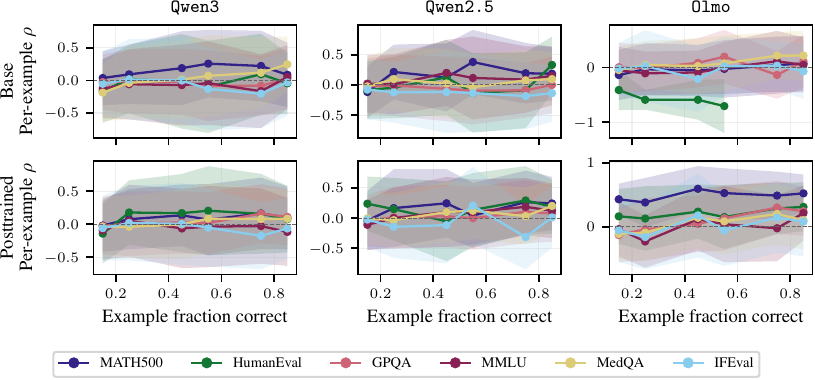}
    \caption{
    \textbf{The more correct a sample is under repeated sampling, the stronger its correlation coefficient.}
    We observe similar results across all model families (columns) and variants (rows).
    Same as \Cref{fig:correlation-vs-fraction-correct-across-methods-model-family-variant} but excluding beam search.
    }
    \label{fig:correlation-vs-fraction-correct-across-methods-model-family-variant-no-beam}
\end{figure}

\clearpage
\subsection{Log-Probabilities}
\label{app:sec:sequence-level}

In this section, we first analyze sequence \logprobs{} in \Cref{app:sec:sub:global-log-probs}.
Then, we focus on token-level \logprobs{} in \Cref{app:sec:sub:token-level-log-probs}, and investigate the difference between \logprob{} and per-token \logprob{} in \Cref{app:sec:subsub:per-token-log-prob}.

\subsubsection{Log-Probabilities Across Methods}
\label{app:sec:sub:global-log-probs}

In this section, we would like to understand how \logprobs{} differ across local and global methods.
First, in \Cref{fig:correlation-accuracy-by-method-dataset}, we are interested in global trends.

\begin{figure}[h]
    \includegraphics{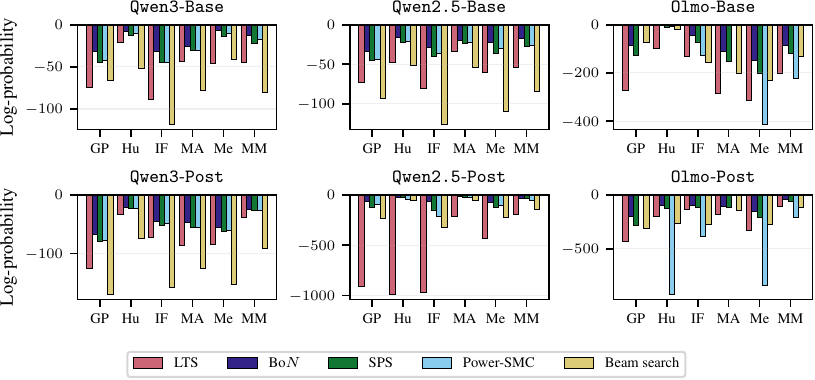}
    \caption{
    \textbf{Beam search often achieves smallest \logprobs{}; Bo$N$, SPS, and power-SMC have larger log-probabilities.}
    Log-probabilities averaged over model sizes.
    Typically, Bo$N$, SPS, and power-SMC lead to smallest log-probabilities.
    LTS is typically larger and beam search is largest.
    \olmo{} seems to be an outlier for power-SMC where \logprobs{} appear very small.
    Datasets are plotted horizontally on the bottom of each panel:
    \underline{GP}QA,
    \underline{Hu}maneval,
    \underline{IF}Eval,
    \underline{MA}TH500,
    \underline{Me}dQA,
    \underline{MM}LU.
    }
    \label{fig:correlation-accuracy-by-method-dataset}
\end{figure}

We find that there are various noticeable differences.
Typically, all global methods achieve similarly large log-probabilities across models and datasets.
Beam search seems to be an outlier reaching significantly lower log-probabilities, especially for \qwenthree{} and \qwentwofive{}. 
We attribute this to the larger sequence lengths (see also \Cref{app:sec:sub:sequence-lengths}).
LTS reaches similarly, much smaller log-probabilities.
On \olmo{}, we find power-SMC to behave like an outlier.
Typically, Bo$N$ reaches the most probable sequences.

\clearpage
\subsubsection{Token-Level Log-Probabilities}
\label{app:sec:sub:token-level-log-probs}

Next, we compare LTS, Bo$N$, and SPS in \Cref{fig:token-position-vs-per-token-log-prob} and \Cref{fig:token-position-vs-per-token-log-prob-1.7b} on a token-level.
We plot per-token-\logprob{} in the left panels and the change of token-\logprob{} in the right panels for base \qwenthree{-8B-Base} (top) and \qwenthree{-8B} (bottom) on MATH500.
\Cref{fig:token-position-vs-per-token-log-prob-1.7b} shows the same plot for \qwenthree{-1.7B}

\begin{figure}[h]
    \includegraphics{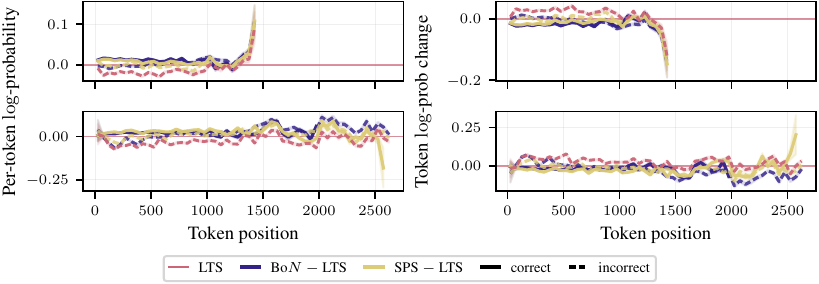}
    \caption{
    \textbf{Global decoding methods produce sequences of larger per-token \logprob{} that are smoother in \texttt{posttrained} models.}
    For \texttt{base} models, LTS samples are shortest (top row), SPS and Bo$N$ samples appear longer.
    There is no meaningful difference between \logprob{} and smoothness.
    For \texttt{posttrained} models, we observe larger per-token log-probability and smoother trajectories for global decoding methods.
    \qwenthree{-8B-Base} (top) and \qwenthree{-8B} (bottom).
    Details in \Cref{app:sec:sequence-level}.
    }
    \label{fig:token-position-vs-per-token-log-prob}
\end{figure}

\begin{figure}[h]
    \includegraphics{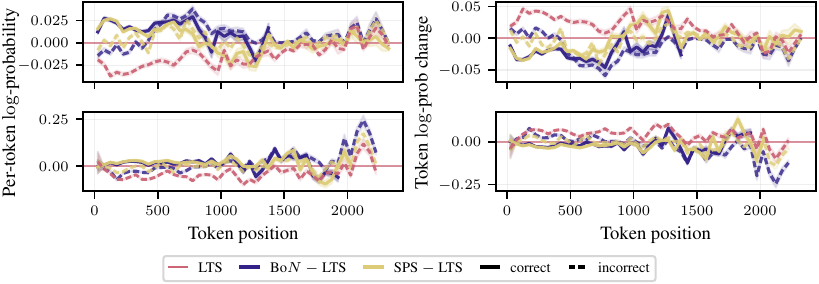}
    \caption{
    \textbf{Global decoding methods produce sequences of larger per-token \logprob{} that are smoother in \texttt{posttrained} models.}
    For \texttt{base} models, LTS samples are shortest (top row), SPS and Bo$N$ samples appear longer.
    There is no meaningful difference between \logprob{} and smoothness.
    For \texttt{posttrained} models, we observe larger per-token log-probability and smoother trajectories for global decoding methods.
    \qwenthree{-1.7B-Base} (top) and \qwenthree{-1.7B} (bottom).
    Details in \Cref{app:sec:sequence-level}.
    }
    \label{fig:token-position-vs-per-token-log-prob-1.7b}
\end{figure}

For \texttt{base} models, we find that samples are shorter than for \texttt{posttrained} models.
Log-probabilities are smaller for incorrect sequences from LTS.
There is no real difference in \logprob{} and smoothness comparing Bo$N$ and SPS.
Generally, we find that per-token \logprobs{} for global methods are larger than LTS. 
Similarly, global methods seem to produce sequences that are more smooth.
However, we note that these effects seem to be rather small and sometimes even inconsistent.

\clearpage
\subsubsection{Log-Probability and Per-Token Log-Probability}
\label{app:sec:subsub:per-token-log-prob}

Here, we investigate the difference of \logprob{} and per-token \logprob{} (averaged over the sequence lengths) on a dataset level, within and across methods.

\paragraph{Within-dataset correlation.}
We investigate how our results are impacted by \logprob{} instead of per-token average \logprob{} (i.e., multiplying \logprob{} by the inverse sequence length).
In \Cref{fig:sum-vs-average-log-prob} we find that using per-token \logprob{} does not change the within-dataset correlation significantly when considering Spearman $\rho$.
For Pearson $r$, many points get pushed to the corner;
the off-diagonal elements seem to be consistent for $\rho$ and $r$.

\begin{figure}[h]
    \includegraphics{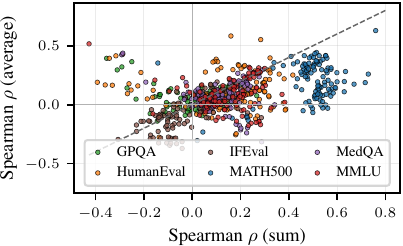}
    \includegraphics{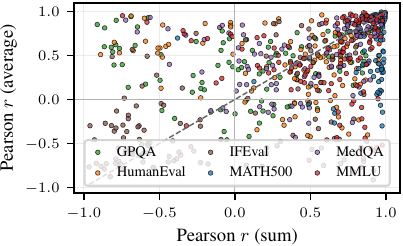}
    \caption{
    \textbf{With Spearman $\rho$ there is no significant difference between \logprob{} and per-token \logprob{} while Pearson $r$ pushes points to the corners.}
    Details in \Cref{app:sec:sequence-level}.
    }
    \label{fig:sum-vs-average-log-prob}
\end{figure}

\begin{figure}[h]
    \includegraphics{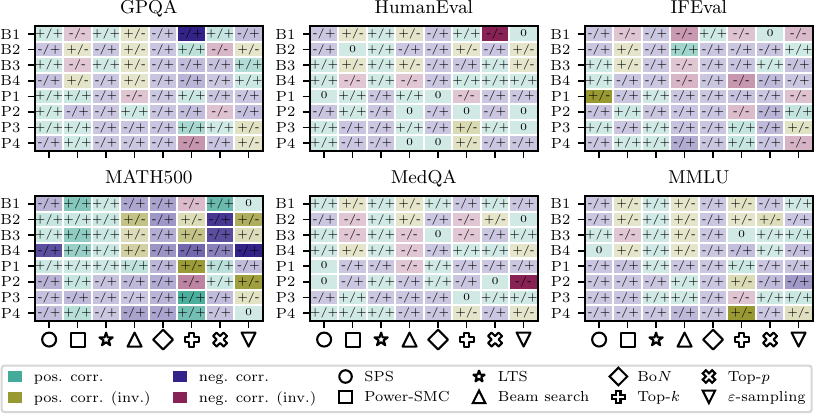}
    \caption{
    \textbf{Methods show either positive or negative correlation within their hyperparameter using per-token \logprob{}.}
    Correlations within methods for local and global decoding methods, models (\qwenthree{} series $0.6\mathrm B$ as $\ast$1, $1.7\mathrm B$ as $\ast$2, $4\mathrm B$ as $\ast$3, and $8\mathrm B$ as $\ast$4 with base as B$\ast$ and posttrained model as P$\ast$), and benchmark datasets.
    Correlations are not consistent across datasets and models.
    }
    \label{fig:correlation-within-method-qwen3-avg}
\end{figure}

\begin{figure}[h]
    \includegraphics{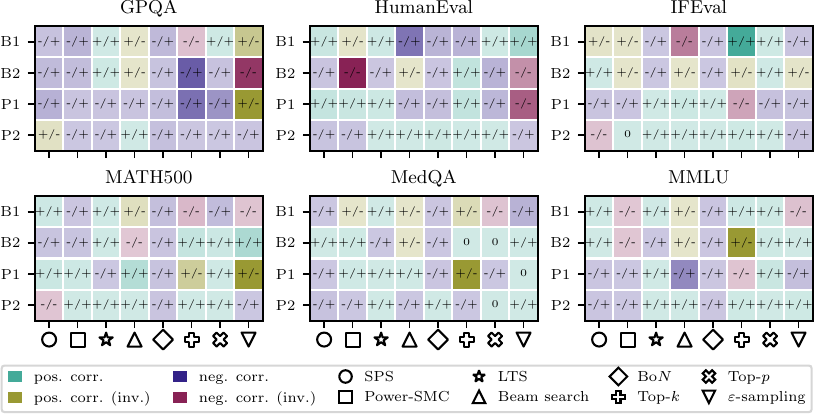}
    \caption{
    \textbf{Methods show either positive or negative correlation within their hyperparameter using per-token \logprob{}.}
    Correlations within methods for local and global decoding methods, models (\qwentwofive{} series $8\mathrm B$ as $\ast$1, \texttt{Math} $8\mathrm B$ as $\ast$2 with base as B$\ast$ and posttrained model as P$\ast$), and benchmark datasets.
    Correlations are not consistent across datasets and models.
    }
    \label{fig:correlation-within-method-qwen25-avg}
\end{figure}

\begin{figure}[h]
    \includegraphics{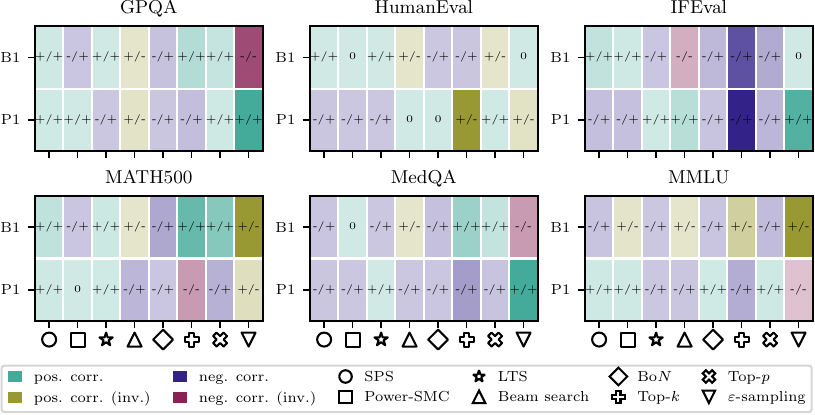}
    \caption{
    \textbf{Methods show either positive or negative correlation within their hyperparameter using per-token \logprob{}.}
    Correlations within methods for local and global decoding methods, models (\olmo{} series $7\mathrm B$ as $\ast$1 with base as B$\ast$ and posttrained model as P$\ast$), and benchmark datasets.
    Correlations are not consistent across datasets and models.
    }
    \label{fig:correlation-within-method-olmo-avg}
\end{figure}

\paragraph{Within-method correlations.}
Considering \Cref{fig:correlation-within-method-qwen3-avg}, \Cref{fig:correlation-within-method-qwen25-avg}, and \Cref{fig:correlation-within-method-olmo-avg}, we find that mostly beam search seems to flip from negative correlation to positive correlation (inverse).
Sometimes also power-SMC flipps a negative correlation.
Other methods seem to not change significantly.

\paragraph{Across-method correlations.}
Across methods, it seems like many originally yellow cells ($+$/$-$) flip to red ($-$/$-$) for SPS and power-SMC aligning with the LTS implied baseline.
Similarly, many of the blue cells ($-$/$+$) flip to green.
For beam search, we find many yellow cells ($+$/$-$) flipping to gree cells; however, many red cells ($-$/$-$) flip to yellow ($+$/$-$).
The resulting plot still does not shows a consistent direction, verifying the claims from the main text.
While some of the cells flip towards more correlation, some flip to less correlation.

\begin{figure}[h]
    \includegraphics{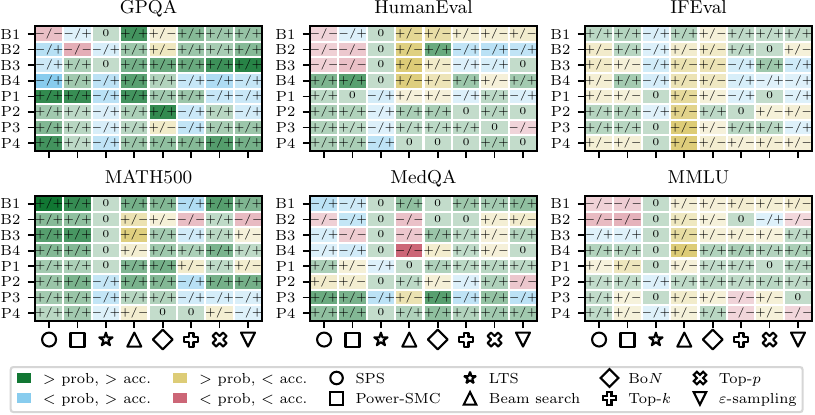}
    \caption{
    \textbf{Various datasets and methods show positive correlations between sequence probability and correctness using per-token \logprob{}.}
    Correlations across methods for local and global decoding methods, models (\qwenthree{} series $0.6\mathrm B$ as $\ast$1, $1.7\mathrm B$ as $\ast$2, $4\mathrm B$ as $\ast$3, and $8\mathrm B$ as $\ast$4 with base as B$\ast$ and posttrained model as P$\ast$), and benchmark datasets.
    Correlations are not consistent across datasets and models.
    }
    \label{fig:correlation-accuracy-by-method-dataset-qwen3-avg}
\end{figure}

\begin{figure}[h]
    \includegraphics{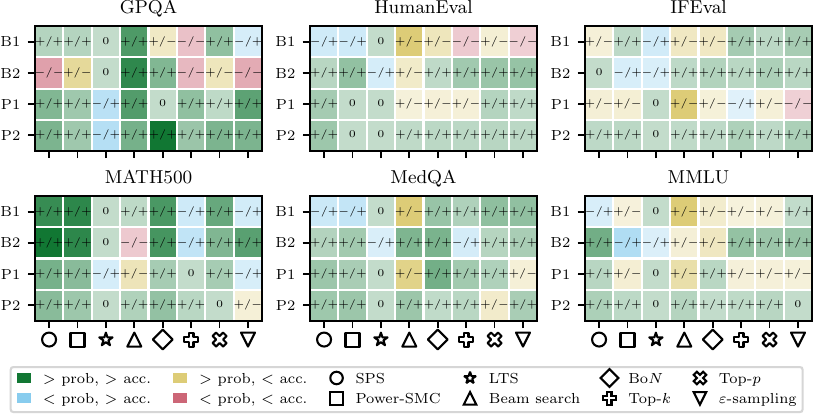}
    \caption{
    \textbf{Various datasets and methods show positive correlations between sequence probability and correctness using per-token \logprob{}.}
    Correlations across methods for local and global decoding methods, models (\qwentwofive{} series $8\mathrm B$ as $\ast$1, \texttt{Math} $8\mathrm B$ as $\ast$2 with base as B$\ast$ and posttrained model as P$\ast$), and benchmark datasets.
    Correlations are not consistent across datasets and models.
    }
    \label{fig:correlation-accuracy-by-method-dataset-qwen25-avg}
\end{figure}

\begin{figure}[h]
    \includegraphics{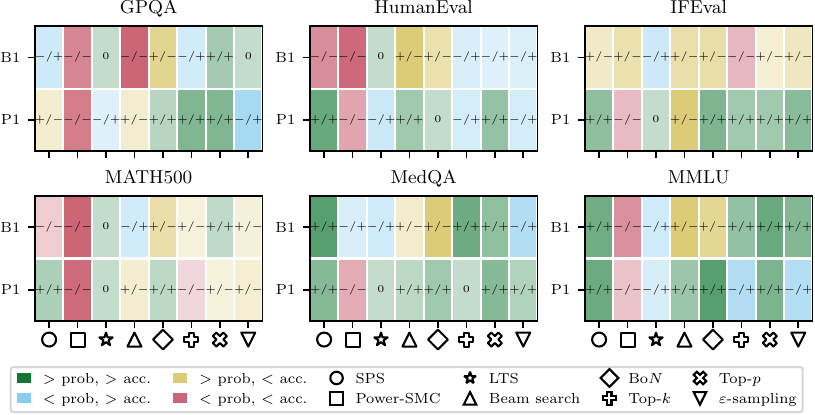}
    \caption{
    \textbf{Various datasets and methods show positive correlations between sequence probability and correctness using per-token \logprob{}.}
    Correlations across methods for local and global decoding methods, models(\olmo{} series $7\mathrm B$ as $\ast$1 with base as B$\ast$ and posttrained model as P$\ast$), and benchmark datasets.
    Correlations are not consistent across datasets and models.
    }
    \label{fig:correlation-accuracy-by-method-dataset-olmo-avg}
\end{figure}

\clearpage
\subsection{Sequence Lengths}
\label{app:sec:sub:sequence-lengths}

\Cref{fig:sequence-length-base} and \Cref{fig:sequence-length-post} show \qwenthree{-8B} \texttt{base} and \texttt{posttrained} models, respectively.
We find that the sequence lenghts are very similar across methods. 
Beam search appears to be the only exception finding very long sequences.

\begin{figure}[h]
    \includegraphics{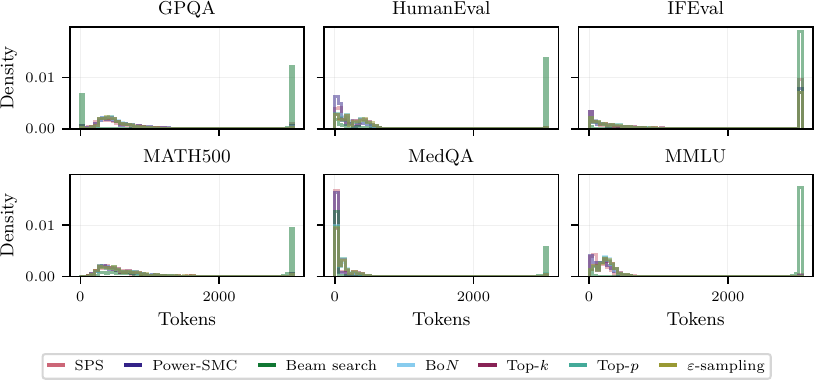}
    \caption{
    \textbf{Sequence lengths overlap across methods, beam search is the only exception.}
    Data for \qwenthree{-8B-Base}.
    Details in \Cref{app:sec:sequence-level}.
    }
    \label{fig:sequence-length-base}
\end{figure}

\begin{figure}[h]
    \includegraphics{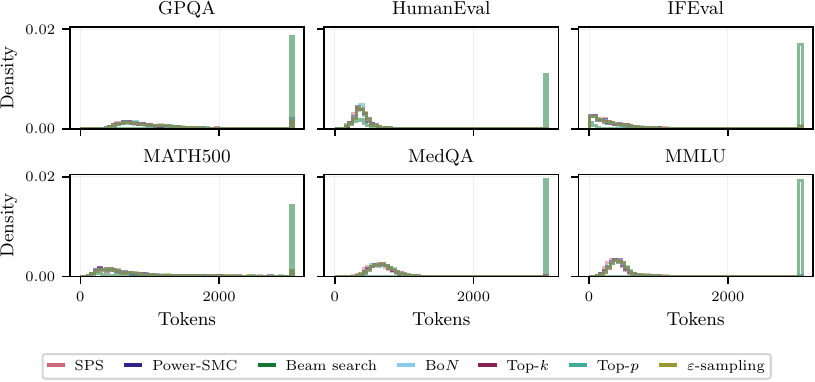}
    \caption{
    \textbf{Sequence lengths overlap across methods, beam search is the only exception.}
    Data for \qwenthree{-8B}.
    Details in \Cref{app:sec:sequence-level}.
    }
    \label{fig:sequence-length-post}
\end{figure}

\clearpage
\subsection{Mode and Power Distribution}
\label{sec:sub:mode-vs-power-distribution}

In \Cref{sec:sampling-methods}, we found that the power-distribution can be understood as a soft mode-seeking objective.
Here, we compare both distributions using Bo$N$ (which returns the mode for $N\to\infty$) for the mode and SPS and power-SMC for the power distribution.

\begin{figure}[h]
    \includegraphics{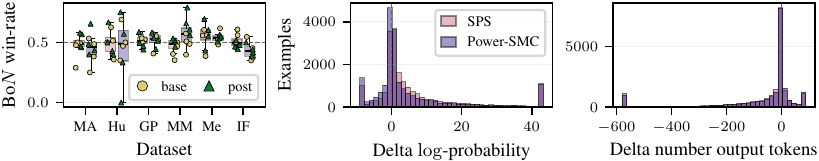}
    \caption{
    Left: \textbf{Comparing mode and power-distribution we find no consistent differences in accuracy.}
    Middle, right: \textbf{Bo$N$ typically leads to shorter samples that are of smaller probability.}
    Datasets are plotted horizontally on the bottom of each panel:
    \underline{GP}QA,
    \underline{Hu}maneval,
    \underline{IF}Eval,
    \underline{MA}TH500,
    \underline{Me}dQA,
    \underline{MM}LU.
    \qwenthree{} model family.
    }
    \label{fig:accuracy-bon-power-distribution}
\end{figure}

Comparing correctness across Bo$N$ and power-distribution in \Cref{fig:accuracy-bon-power-distribution} (leftmost panel), we find no consistent trend across datasets and models (and variants).
In the panel, we plot the win-rate of Bo$N$ against SPS and power-SMC across datasets and the \qwenthree{} series.
Investigating the samples in detail, we find that samples from Bo$N$ seem to be typically of smaller probability (\Cref{fig:accuracy-bon-power-distribution}, middle) and shorter, two properties that influence each other.

\clearpage
\subsection{Self-Distillation}
\label{app:sec:sub:self-distillation}

\citet{hinton2015distilling} distill an ensemble of neural networks in a single neural network.
\citet{zhang2019your} propose self-distillation for convolutional neural networks where knowledge is distilled within a network itself.
In the context of LLMs, the self-taught reasoner \citep{zelikman2022star} consists of a bootstrapping method that lets LLMs generate rationales for prompts which are then used for finetuning.
\citet{chen2024language} frames reasoning as sampling from a latent distribution optimizing it via sequence likelihood as self-rewarding signal.

Here, we train self-distillation on Bo$N$, SPS, and power-SMC samples using the first half of samples considered.
We test the model consistently on the second half of samples.

For fine-tuning we choose a LoRA with $r=16$, $\alpha=32$ and dropout of $0.05$ for $1$ epoch with a learning rate of $2\cdot10^{-5}$ and a warmup ratio of $0.03$.

\begin{figure}[h]
    \centering
    \includegraphics{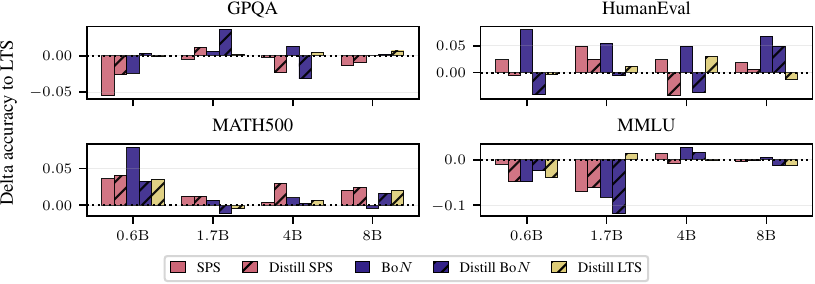}
    \caption{
    \textbf{Self-distillation shows diverse results across methods.}
    On MATH500 power self-distillation seems to often improve results.
    Else, results are mixed.
    \qwenthree{} \texttt{base} models.
    }
    \label{fig:self-distillation}
\end{figure}

In \Cref{fig:self-distillation} we find that self-distillation seems to work well on samples where a correlation between \logprob{} and correctness can be found (i.e., MATH500). 
Else we observe mixed results.
Further, self-distillation on samples from the power distribution works best where those samples are outperforming the baseline LTS (MATH500).
Else, simple distillation on LTS samples often outperforms power samples.

\clearpage
\subsection{\qwenthree{} Thinking}
\label{app:sec:sub:qwen3-thinking}

In this section, we enable thinking for the \qwenthree{} model series but keep the maximum number of tokens as before. 

\begin{figure}[h]
    \centering
    \includegraphics{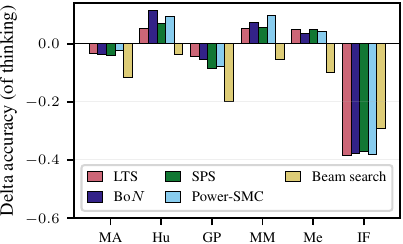}
    \includegraphics{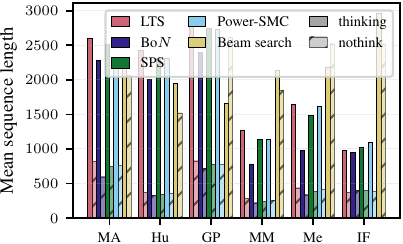}
    \caption{
    Left: \textbf{\qwenthree{} thinking mostly improves on Humaneval, MMLU, and MedQA; the strongest decline can be observed on IFEval.
    }
    Across methods, beam search often performs worst.
    Right: \textbf{Thinking models often reach the token limit.}
    The mean sequence length for thinking models is much higher as compared to non-thinking models.
    }
    \label{fig:qwen3-thinking-summary}
\end{figure}

In \Cref{fig:qwen3-thinking-summary} (left) we find that \qwenthree{} thinking models often degrade the performance ($3$ out of $6$ datasets) across methods (with beam search acting as an exception).
We explain this mostly by the fact that they reach the token limit (due to producing many tokens inside the \texttt{<think>}-tags) and do not come to an answer (\Cref{fig:qwen3-thinking-summary}, right).
As the token limit is dictated by our computational constraints, we focus in the main text solely on the non-thinking version.
It seems interesting to rerun our experiments with thinking models in future work.

\end{document}